\documentclass[twoside,11pt]{article}
\usepackage[accepted]{melba}
\usepackage{stmaryrd}
\usepackage{bbm}
\usepackage{graphicx}
\usepackage{amsmath}
\usepackage{amssymb}
\usepackage{verbatim}
\usepackage[bottom]{footmisc}
\usepackage{graphicx}
\usepackage{soul}
\usepackage{subcaption}
\usepackage{import}
\usepackage{xifthen}
\usepackage{xcolor}
\usepackage{pdfpages}
\usepackage{transparent}
\usepackage{tikz}
\usepackage{caption}
\usepackage{float}

\melbaheading{013}{https://www.melba-journal.org/article/27655}{2021}{1-39}{01/2021}{09/2021}{Camarasa, Bos, Hendrikse, Nederkoorn, Kooi, van der Lugt and de Bruijne}{Uncertainty for Safe Utilization of Machine Learning in Medical Imaging (UNSURE) 2020}{Christian Baumgartner, Adrian Dalca, Carole Sudre, Ryutaro Tanno, Sandy Wells}
\ShortHeadings{Comparison of Epistemic Uncertainty Maps}{Camarasa et al.}
\firstpageno{1}

\begin{document}
\title{
    A Quantitative Comparison of Epistemic Uncertainty Maps Applied to Multi-Class Segmentation
}

\author{\name Robin Camarasa \email r.camarasa@erasmusmc.nl\\
    \addr Biomedical Imaging Group Rotterdam, Department of Radiology and Nuclear Medicine, Erasmus MC, Rotterdam, The Netherlands
    \AND
    \name Daniel Bos  \email d.bos@erasmusmc.nl\\
    \addr Department of Radiology and Nuclear Medicine, Erasmus MC, Rotterdam, The Netherlands\\
    \addr Department of Epidemiology, Erasmus MC, Rotterdam, The Netherlands
    \AND
    \name Jeroen Hendrikse  \email j.hendrikse@umcutrecht.nl\\
    \addr Department of Radiology, University Medical Center Utrecht, Utrecht, The Netherlands
    \AND
    \name Paul Nederkoorn  \email p.j.nederkoorn@amsterdamumc.nl\\
    \addr Department of Neurology, Academic Medical Center University of Amsterdam, Amsterdam, The Netherlands
    \AND
    \name M. Eline Kooi  \email eline.kooi@mumc.nl\\
    \addr Department of Radiology and Nuclear Medecine, CARIM School for Cardiovascular Diseases, Maastricht University Medical Center, Maastricht, The Netherlands
    \AND
    \name Aad van der Lugt  \email a.vanderlugt@erasmusmc.nl\\
    \addr Department of Radiology and Nuclear Medicine, Erasmus MC, Rotterdam, The Netherlands
    \AND
    \name Marleen de Bruijne  \email marleen.debruijne@erasmusmc.nl\\
    \addr Biomedical Imaging Group Rotterdam, Department of Radiology and Nuclear Medicine, Erasmus MC, Rotterdam, The Netherlands\\
    \addr Department of Computer Science, University of Copenhagen, Denmark
    \AND
}

\maketitle

\vspace{-5em}
\begin{abstract}
    Uncertainty assessment has gained rapid interest in medical image analysis. A popular technique to compute epistemic uncertainty is the Monte-Carlo (MC) dropout technique. From a network with MC dropout and a single input, multiple outputs can be sampled. Various methods can be used to obtain epistemic uncertainty maps from those multiple outputs. In the case of multi-class segmentation, the number of methods is even larger as epistemic uncertainty can be computed voxelwise per class or voxelwise per image. 

    This paper highlights a systematic approach to define and quantitatively compare those methods in two different contexts: class-specific epistemic uncertainty maps (one value per image, voxel and class) and combined epistemic uncertainty maps (one value per image and voxel). We applied this quantitative analysis to a multi-class segmentation of the carotid artery lumen and vessel wall, on a multi-center, multi-scanner, multi-sequence dataset of Magnetic Resonance (MR) images. We validated our analysis over 144 sets of hyperparameters of a model. 

    Our main analysis considers the relationship between the order of the voxels sorted according to their epistemic uncertainty values and the misclassification of the prediction. Under this consideration, the comparison of combined uncertainty maps reveals that the multi-class entropy and the multi-class mutual information statistically out-perform the other combined uncertainty maps under study (the averaged entropy, the averaged variance, the similarity Bhattacharya coefficient and the similarity Kullback-Leibler divergence). In a class-specific scenario, the one-versus-all entropy statistically out-performs the class-wise entropy, the class-wise variance and the one versus all mutual information. The class-wise entropy statistically out-performs the other class-specific uncertainty maps in term of calibration. We made a python package available to reproduce our analysis on different data and tasks. 
\end{abstract} 


\section{Introduction}

Deep learning has become an important asset in medical image segmentation due to its good performance \citep{taghanaki2021deep}. However, a major obstacle to its application in clinical practice is the lack of uncertainty assessment; without this it is impossible to know how trustworthy the prediction of a deep learning algorithm is. \cite{kiureghian2009aleatory} discriminates uncertainty into: epistemic uncertainty and aleatoric uncertainty. Epistemic uncertainty is inherent to the model and decreases with an increased amount of data, while aleatoric uncertainty is inherent to the input data. In clinical practice, epistemic uncertainty would indicate the degree of uncertainty of the algorithm regarding its prediction, and the aleatoric uncertainty of the algorithm regarding the quality of the input data (for example: artefacts, blurring...).

Bayesian deep learning provides a mathematically grounded answer to epistemic uncertainty assessment and was therefore investigated since the rise of computer vision \citep{mackay1992practical, denker1991transforming, neal1993bayesian}. These methods allow epistemic uncertainty assessment since the neurons of a Bayesian neural network contain distributions instead of scalars (distributions from which it is possible to extract uncertainties). However early methods substantially increased the time and space complexity of those algorithms compared to their non-Bayesian counterparts. The work of \cite{gal2016dropout} renewed the interest for the field as they developed the MC dropout framework: a form of Bayesian inference that requires only a slight alteration of the current state of the art models and conserves the time complexity of standard approaches.

 For segmentation tasks, uncertainty assessment can be expressed in the form of uncertainty maps specifying the uncertainty for each voxel of a given image. \cite{nair2020exploring} analysed both aleatoric and epistemic uncertainty maps in a single-class segmentation task:segmenting brain lesion on MR images. Considering each class separately, \cite{mehta2020uncertainty} developed a performance measure to compare different uncertainty quantification methods on a multi-class segmentation task. However, a multi-class segmentation task provide a larger set of epistemic uncertainty maps than a single-class segmentation task as it is possible to compute combined epistemic uncertainty maps (one uncertainty map per voxel, per image) or class-specific epistemic uncertainty maps (one uncertainty map per class, per voxel, per image). Both options are clinically relevant as the combined epistemic uncertainty maps provide a summary of an uncertainty of the algorithm of all the segmented classes simultaneously where a class-specific epistemic uncertainty maps considers the segmented classes separately. For both combined epistemic uncertainty maps and class-specific epistemic uncertainty maps many design choice are possible,however there is limited research focusses on analysing the extra options resulting from the multi-class setting.

 Carotid atherosclerosis, corresponding to the thickening of the vessel wall of the carotid artery, is one of the most important risk factor for ischemic stroke \citep{world2011global}, which in turn is an important cause of death and disability worldwide \citep{world2014global}. Carotid atherosclerosis can be visualized in vivo through non-invasive modalities such as Ultrasound, Computed Tomography or Magnetic Resonance Imaging and can be assessed measuring the degree of stenosis (ratio of the diameter of the vessel wall and of the diameter of the lumen). This measurement can be obtained from the segmentation of the lumen and vessel wall of the carotid artery. Therefore a trustworthy segmentation of the carotid artery is an important step towards early detection of an increased risk of ischemic strokes.

This article extends our work presented at the MICCAI UNSURE workshop \citep{camarasa2020quantitative}. In addition to the findings of our previous work that only considered combined epistemic uncertainty maps, this paper provides an approach to characterise different class-specific epistemic uncertainty maps, combined epistemic uncertainty maps and analyses additional measures of uncertainty. 

In this work, our contribution is threefold. Firstly, we provide a systematic approach to characterize both class-specific epistemic uncertainty maps, combined epistemic uncertainty maps separating the epistemic uncertainty map into a uncertainty measure and an aggregation method. Secondly, we quantitatively and statistically compare the ability of those different epistemic uncertainty maps to assess misclassification on a segmentation of the carotid artery on a multi-center, multi-scanner, multi-sequence dataset of MR images. However, this analysis is not specific to this task. Thirdly, we compare our evaluation of class-specific epistemic uncertainty derived from the MC dropout technique to the one proposed in the BRATS challenge \citep{menze2014multimodal} in a multi-class segmentation setting. We provide a python package\footnote{pypi url: \url{https://pypi.org/project/monte-carlo-analysis} \newline GitLab url: \url{https://gitlab.com/python-packages2/monte-carlo-analysis}} available on the pypi platform to facilitate reproduction of this analysis on other data and tasks.

\section{Related work}

\subsection{Carotid artery segmentation}

Automatic segmentation of lumen and vessel wall of the carotid artery on MR sequences is a rising field. \cite{luo2019carotid} developed a segmentation method on TOF-MRA images based on level set method. \cite{arias2018maximization} used an optimal surface graph-cut algorithm to segment the lumen and vessel wall of different MR sequences. \cite{wu2019deep} compared different state-of-the-art deep learning methods to segment lumen and vessel wall of the carotid artery on 2D T1-w MR images. \cite{zhu2021cascaded} segmented the carotid artery on a multi-sequence dataset using a cascaded 3D residual U-net.

\subsection{Uncertainty in deep learning}

Uncertainties are commonly divided in epistemic and aleatoric uncertainties as detailed by \cite{kiureghian2009aleatory}. Epistemic uncertainty is inherent to the model and decreases with increasing dataset size, while aleatoric uncertainty is inherent to the data and corresponds to the randomness of the input data. Aleatoric uncertainties can be further subdivided into homoscedastic uncertainties (aleatoric uncertainties that are the same for all inputs) and heteroscedastic uncertainties (aleatoric uncertainties specific to each input). Finally, distributional uncertainties are described as the uncertainty due to the differences between training and test data (also referred to as data shift).

Various approaches can be found in the literature to estimate the aleatoric uncertainties. \cite{kendall2017uncertainties} proposed to learn those uncertainties by modifying the loss and \cite{wang2019aleatoric} proposed to generate images from the input space using test-time augmentation. To assess epistemic uncertainty, Bayesian Deep Learning methods offer a well grounded mathematical approach. Usually, the prediction of those methods rely on Bayesian model averaging which consists of a marginalisation over a distribution on the weights of the model; the epistemic uncertainty is also derived from this distribution on the weights of the model \citep{wilson2020bayesian}. \cite{blundell2015weight} proposed the methodology Bayes by backpropagation, to learn this distribution on the weights. Another popular approach is to sample a set of weights using stochastic gradient Langevin dynamics \citep{welling2011bayesian, izmailov2018averaging, maddox2019simple}. Alternatively, \cite{lakshminarayanan2016simple} used deep ensembles to obtain multiple predictions. While the authors claimed their approach is non-Bayesian, \cite{wilson2020bayesian} argued that deep ensembles fall within the category of Bayesian model averaging methods. Many recent papers applied Bayesian deep learning methods to medical segmentation tasks as highlighted in the review of uncertainty quantification in deep learning by \citep{abdar2020review}.

\subsection{Monte-Carlo dropout}

A widely used Bayesian method to measure epistemic uncertainties is Monte-Carlo dropout \citep{gal2016dropout}. Monte-Carlo dropout applies dropout at both training and test time. The Bayesian model averaging is then performed by sampling different set of weights using dropout at test time. \cite{mukhoti2018evaluating} proposed the method "concrete dropout" to tune the dropout rate of the Monte-Carlo dropout technique. Other stochastic regularisation techniques were also investigated in the literature as an alternative for Monte-Carlo dropout such as Monte-Carlo DropConnect \citep{mobiny2019dropconnect}, Monte-Carlo Batch Normalization \citep{teye2018bayesian}. In medical image analysis, Monte-Carlo dropout found  application in many tasks including classification \citep{mobiny2019risk}, segmentation \citep{orlando2019u2}, regression \citep{laves2020well} and registration \citep{sedghi2019probabilistic}.

\subsection{Uncertainty analysis}

Analysing qualitatively and quantitatively  uncertainties is an active research question. To evaluate distributional uncertainties, \cite{blum2019fishyscapes} introduced a datashift in the test-set and compared the ability of different distributional uncertainty methods to detect this datashift on the Cityscapes dataset. In the field of computer vision, \cite{gustafsson2020evaluating} compared the ability of Monte-Carlo dropout and Deep-ensemble methods to evaluate ordering and calibration of epistemic uncertainties. In medical imaging, a straightforward approach is to consider the inter-observer variability as "ground truth" aleatoric uncertainties. \cite{jungo2018effect} studied the relationship between the "ground truth" aleatoric uncertainties and the epistemic uncertainties of a MC dropout model evaluating the weighted mean entropy. \cite{chotzoglou2019exploring} compared predicted aleatoric uncertainties to "ground truth" aleatoric uncertainties using ROC curves. Alternatively, in a skin lesion classification problem, \cite{de2019quantifying} introduced an uncertainty measure based on distribution similarity of the two most probable classes. The authors recommend the use of this uncertainty measure compared to variance based ones since it is more interpretable as the range of uncertainty values is between 0 and 1 (0 very certain, 1 very uncertain). In another work, \cite{mehrtash2019confidence} compared calibrated and uncalibrated segmentation with negative log likelihood and Brier score. \cite{jungo2020analyzing} compared different aleatoric uncertainty maps and epistemic uncertainty maps based on different Bayesian model averaging methods using the Dice coefficient between the uncertainty map and the misclassification. Finally, \cite{nair2020exploring} assessed both aleatoric and epistemic uncertainties via the relation to segmentation errors. To study this relationship they compare the gain in performance when filtering out the most uncertain voxels for different epistemic and aleatoric uncertainty maps in a single-class segmentation problem.

\section{Methods}

\begin{figure}[t]
    \centering
    \def\svgwidth{\columnwidth}
    \tiny 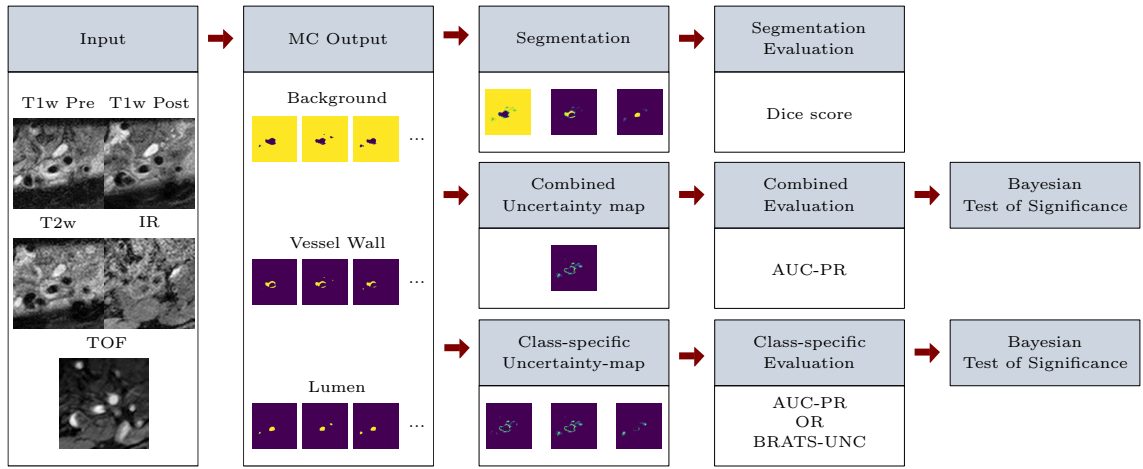
    \caption{Study overview}
    \label{fig:pipeline}
\end{figure}

We compare epistemic uncertainty maps of carotid artery segmentation. The epistemic uncertainty maps are based on the MC Dropout technique described in Subsection \ref{mcdropout}. The different uncertainty maps are derived from the outputs of MC dropout using different measures of uncertainty and aggregation methods, respectively detailed in Subsections \ref{uncertaintymetric} and \ref{aggregations}. The method to assess the quality of the uncertainty maps and to assess the significance of the comparison of uncertainty maps are described in Subsections \ref{evaluation} and \ref{statisticalsignificance} respectively. An overview of this method is provided in Figure \ref{fig:pipeline}.

\subsection{MC-dropout} \label{mcdropout}

The MC Dropout technique consists of using dropout both at training and testing time. While adding dropout is the only alteration of the training procedure, the testing procedure requires the evaluation of multiple outputs at test time for a single input. From these multiple outputs, one can obtain a prediction and class-specific or combined epistemic uncertainty maps that will be detailed later on (Section \ref{uncertaintymetric} and \ref{aggregations}).

\sloppy In the following, $(n_x, n_y, n_z) \in \mathbb{N}^3$ represents the dimensions of the input images, $K$ the number of input sequences, $C$ the number of output classes, $x \in \mathbb{R}^{n_x \times n_y \times n_z \times K}$ an input image $x$, $j \in J=\{0, ..., n_x-1\} \times \{0, ..., n_y-1\} \times \{0, ..., n_z-1\}$ a 3D coordinate. $\theta \in \Omega$ represents the parameters (weights and biases) of a trained segmentation network. We define the following distributions: 
\begin{enumerate}
    \item $y_{j}$ represents the prediction distribution for a given input $x$ and a given voxel $j$; this distribution has the following support: $\text{supp}(y_j) = \{(z^c)_{1 \leq c \leq C} \in [0,1]^C | \sum_{c=1}^C z_c = 1\}$    
    \item$y_{j}^c$ represents the class-specific prediction distribution for given input $x$ a given class $c$ and a voxel $j$; this distribution has the following support $\text{supp}(y^c_j) = [0, 1]$
    \item $Y_j$ represents the MC dropout distribution for a given input $x$; this distribution has the following support $\text{supp}(Y_j) = \{1, ..., C\}$
\end{enumerate}

In practice, $y_j$ and $y_j^c$ correspond to the distribution over the output before applying the Bayesian model averaging for all classes and a given class $c$ respectively. $Y_j$ correspond to the distribution over the output after applying the Bayesian model averaging. Note that $y_j$ and $y_j^c$ have a continuous support where $Y_j$ have a discrete support.

In segmentation using MC dropout \citep{gal2016dropout} - to obtain several estimates of the (multi-class) segmentation - we sample $T$ sets of parameters $(\theta_1, ..., \theta_T)$ dropping out feature maps at test time. From those parameters, we can evaluate $T$ outputs per voxel per class $(\text{P}(Y_{j} | \theta = \theta_1), ..., \text{P}(Y_{j} | \theta = \theta_T))$ which represent samples from the prediction distribution $y_j$. From this sample, one can obtain the Bayesian model averaging of the MC dropout ensemble deriving the mean of the class-specific prediction distribution at a voxel level as follows:

\begin{align}
    \text{P}(Y_j = c) = \mathbb{E}(y_j^c)\approx \frac{1}{T} \sum_{t = 1}^T \text{P}(Y_{j}=c | \theta = \theta_t).    
    \label{E:mc_output}
\end{align}

An alternative to the original (Bernoulli) dropout that applies binary noise at a feature map level is to use Gaussian multiplicative noise. Originally this approach was proposed by \cite{srivastava2014dropout} and \cite{gal2016dropout} guaranteed its compatibility with the MC dropout framework. \cite{srivastava2014dropout} proposed the following definition of the Bernouilli dropout and Gaussian multiplicative noise to match their expected mean and variance:

\begin{align}
    \left\{
    \begin{array}{ll}
        B = \lambda A\\
        \lambda_{\text{Bernoulli}} \sim \frac{1}{1 - p} \mathcal{B}(1 - p) \\
        \lambda_{\text{Gaussian}} \sim \mathcal{N}(1, \frac{p}{1 - p})
    \end{array}
    \right.
\label{E:dropout}
\end{align}

where $A$ is a feature map of a dropout layer input, $B$ is the corresponding feature map of that dropout layer output, $\lambda \in \mathbb{R}$ is randomly sampled from the dropout distribution, $p$ is the dropout rate, $\mathcal{B}$ is a Bernoulli distribution and $\mathcal{N}$ is a Gaussian distribution.

\subsection{Quantifying uncertainty}\label{uncertaintymetric}

At test time, MC Dropout technique produces for each voxel $j$ a sample of the prediction distribution $y_j$. To quantify the uncertainty of the prediction distribution per voxel, one can either determine the uncertainty of a class-specific prediction distribution $y_j^c$ as described in subsection \ref{description_metrics}, study the similarity between the different class-specific prediction distributions ($y_j^{c'}$, $y_j^{c''}$) as described in subsection \ref{similarity_metrics} or directly determine the uncertainty of the prediction distribution $y_j$ as described in subsection \ref{multi-class_metrics}. Those three approaches are investigated below, proposing different measures of uncertainty for each approach.

\subsubsection{Description measures} \label{description_metrics}

The first measures investigated are the description measures. Those measures describe the uncertainty of a single class-specific prediction distribution. The use of description measure to assess uncertainty is widely spread in the literature \citep{seebock2019exploiting, kendall2015bayesian}. In this article, two measures of this type are investigated: the distribution variance and the distribution entropy.

A first possibility to compute the uncertainty of a prediction is to compute its variance, yielding the distribution variance measure:
\begin{align}
    D^v(y_{j}^c) = \mathrm{Var}(y_j^c).
    \label{E:Dv}
\end{align}

Another widely used description measure is the entropy of the distribution \citep{wang2019aleatoric}. In contrast with the variance measure, which can be directly computed from data sampled with MC dropout from the class-specific prediction distribution $y_j^c$, it requires the estimation of an integral defined as follows:

\begin{align}
    D^h(y_j^c) = \mathcal{H}(y_j^c) =  -\int_{0}^{1}f_{y_j^c}(v) \text{log}\left(f_{y_j^{c}}(v)\right) \text{dv}
    \label{E:Dh}
\end{align}
where $\mathcal{H}$ is the entropy of a distribution and $f_{y_j^{c}}$ is the probability density function of the class-specific prediction distribution $y_j^c$.

\subsubsection{Similarity measures} \label{similarity_metrics}

A second approach to quantify the uncertainty is to measure the similarity of two class-specific prediction distributions $y_j^{c'}$, $y_j^{c''}$ where $c'$ and $c''$ are two distinct output classes. The more those two distributions overlap, the more similar they are and the more difficult it is to determine the predicted class, which makes the outcome more uncertain. \cite{de2019quantifying} introduced this approach using the Bhattacharya coefficient (BC) to measure uncertainty (0: no overlap, therefore certain; 1: identical, therefore uncertain) as follows:
\begin{align}
    S^b\left(y_j^{c'}, y_j^{c''}\right) = \int_{0}^{1} \sqrt{f_{y_j^{c'}}(v) f_{y_j^{c''}}(v)} \text{dv}.
    \label{E:Sb}
\end{align}

As an alternative measure of distribution similarity, we investigated the Kullback-Leibler divergence. In this case, a high value represents a small overlap among distributions and therefore the negative of the measure is considered. In addition, the Kullback-Leibler (KL) divergence is made symmetric with respect to the distributions $y_j^{c'}$ and $y_j^{c''}$, resulting in:

\begin{align}
    S^{k}\left(y_j^{c'}, y_j^{c''}\right) = -
    \text{KL}\left( y_j^{c'}||y_j^{c''}\right)-
    \text{KL}\left( y_j^{c''}||y_j^{c'}\right)
    \label{E:Sk}
\end{align}

where $\text{KL}$ is the Kullback-Leibler divergence.

\subsubsection{Multi-class measures} \label{multi-class_metrics}

A final option is to capture the uncertainty of the prediction distributions in one measure of uncertainty. A first approach is to use the entropy of the MC dropout distribution (defined as the entropy of the mean of the prediction distribution) \citep{jungo2020analyzing, jungo2018effect, nair2020exploring, mobiny2019risk} as follows:

\begin{align}
    M^h(y_{j}) = \mathcal{H}\left(Y_j\right) = - \sum_{c=1}^C \mathbb{E}\left(y_j^c\right) \text{log}\left(\mathbb{E}\left(y_j^c\right)\right).
    \label{E:Mh}
\end{align}

Note that unlike the distribution entropy of a class-specific prediction distribution computed in Equation \ref{E:Dh}, Equation \ref{E:Mh} does not require the discretisation of an integral.

\cite{michelmore2018evaluating} argue that the mutual information (MI) between the MC dropout distribution $Y_j$ and the distribution over the model parameters $\theta$ captures MC dropout based neural network uncertainty. This approach to compute uncertainty spread in the medical imaging field \citep{nair2020exploring, mobiny2019dropconnect}. One can derive the mutual information between the prediction $Y_j$ and the model parameters $\theta$ by adding a second term to Equation \ref{E:Mh} as in:
\begin{align}
    \begin{split}
        M^m(y_{j})  =& \mathcal{MI}(Y_j, \theta) = \mathcal{H}(Y_j) - \mathcal{H}(Y_j | \theta)\\
                \approx& -\sum_{c=1}^C \mathbb{E}(y_j^c) \text{log}\left(\mathbb{E}(y_j^c)\right) \\
                + & \sum_{c=1}^C \frac{1}{T} \sum_{t=1}^T \text{P}(Y_j=c | \theta=\theta_t) \text{log}\left(\text{P}(Y_j=c | \theta=\theta_t)\right)
    \end{split}
    \label{E:Mm}
\end{align}

\subsection{Aggregation methods}\label{aggregations}

Now that the different uncertainty measures are defined, it is important to define the different aggregation methods to obtain the epistemic uncertainty maps from those measures of uncertainty. Two families of aggregation methods are investigated in this subsection: the combined aggregation methods (one uncertainty map per image) and the class-specific aggregation methods (one uncertainty map per class per image).

\subsubsection{Combined aggregation methods}

 In the case of a combined aggregation method, an intuitive way to assess uncertainty is to use the description measure per voxel per class and average those descriptions per voxel over the classes. This aggregation method, the averaged aggregation method, can be derived as follows:
\begin{align}
    s^{A}_D\left(y_{j}\right) = \frac{1}{C} \sum_{c=1}^C D\left(y_{j}^c\right)
    \label{E:sAD}
\end{align}
where $D$ is a description {measure} (either distribution variance or distribution entropy).

\cite{de2019quantifying} used the similarity of the probability distributions of the two most probable classes to derive an uncertainty measure. This aggregation method referred as the similarity aggregation method can be defined as in:
\begin{align}
    s^{T}_S(y_{j}) = S\left(y_{j}^{c_1}, y_{j}^{c_2}\right)
    \label{E:sTS}
\end{align}
where $c_1$ and $c_2$ are respectively the most and the second most probable classes of the voxel $j$ and $S$ is a similarity measure (either Bhattacharya coefficient or Kullback-Leibler divergence).

A last combined aggregation method investigated in the paper is the multi-class description measure. This aggregation method consists of computing a multi-class measure per voxel as follows:

\begin{align}
    s^{A}_M(y_{j}) = M(y_{j})    
    \label{E:sAM}
\end{align}
where $M$ is a multi-class measure (either entropy or mutual information).

\subsubsection{Class-specific aggregation methods}

The most direct aggregation method to obtain an uncertainty map per class is to compute a description measure (either distribution variance or distribution entropy) per voxel per class. This aggregation method, which we refer to as the description aggregation method, can be described as follows:

\begin{align}
    s^{CD}_{D}\left(y_{j}^{c}\right) = D\left(y_{j}^{c}\right)
    \label{E:sCWD}
\end{align}
where $c'$ is an output class and $D$ a description measure (either distribution variance or distribution entropy).

Alternatively, it is possible to apply a one versus all aggregation method. This aggregation method consists of applying a multi-class measure to the distribution of the class under study and the sum of the distributions of the other classes. This aggregation method allows the use of a multi-class measure to obtain an uncertainty map per class as in: 
\begin{align}
    s^{1vA}_{M}\left(y_{j}^{c}\right) = M\left(\left(y_{j}^c, \overline{y_{j}^c}\right)\right).
    \label{E:s1VA}
\end{align}
where $c'$ is an output class and $M$ is a multi-class measure (either entropy or mutual information).

\subsection{Evaluation}\label{evaluation}

\subsubsection{Combined evaluation}\label{image_wise_evaluation}

For the combined case, \cite{mobiny2019dropconnect} provides a framework to assess the quality of an uncertainty map. Considering uncertainty as a score that predicts misclassification leads to a redefinition of the notions of true and false positives and negatives in an uncertainty context. A voxel is considered misclassified when its predicted class and its ground truth class mismatch. Once an uncertainty map is thresholded at a value $\tau_u$, one can define four types of voxels as summarized in Table \ref{tab:tableuncertainty}: misclassified and uncertain ($\text{UTP}(\tau_u)$ in a sense that the uncertainty of the voxel accurately predicts its misclassification), misclassified and certain ($\text{UFN}(\tau_u)$), correctly classified and uncertain ($\text{UFP}(\tau_u)$) and correctly classified and certain ($\text{UTN}(\tau_u)$). 

For a given value of the uncertainty threshold $\tau_u$, it is possible to compute the precision and the recall of uncertainty as a misclassification predictor following : $\text{UPr}(\tau_u) = \frac{\text{UTP}(\tau_u)}{\text{UTP}(\tau_u) + \text{UFP}(\tau_u)}$ and  $\text{URc}(\tau_u) = \frac{\text{UTP}(\tau_u)}{\text{UTP}(\tau_u) + \text{UFN}(\tau_u)}$. One can compute the area under the precision recall curve (AUC-PR) using scikit-learn implementation \citep{pedregosa2011scikit}.
\begin{table}[t]
    \centering
    \caption{Uncertainty as a predictor of misclassification}
    \begin{tabular}{l|l|l}
    \hline
                            & Uncertain                     & Certain                     \\
    \hline
    Misclassified           & $\text{UTP}(\tau_u)$          & $\text{UFN}(\tau_u)$        \\
    Correctly classified    & $\text{UFP}(\tau_u)$          & $\text{UTN}(\tau_u)$        \\
    \hline
    \end{tabular}
    \label{tab:tableuncertainty}
\end{table}

The main characteristic of this performance measure is its independence from uncertainty map calibration. Only the order of the voxels sorted according to their epistemic uncertainty values matters as this performance measure is invariant by strictly monotonic increasing transformation.

\subsubsection{Class-specific evaluation}

    One can adapt the combined AUC-PR performance measure into a class-specific version. In this scenario, the only misclassified voxels considered are those where the predicted class or the ground truth class is the class under study, all other voxels are considered correctly classified. With this definition of misclassification and an class-specific uncertainty map, one can compute a class-specific version of the AUC-PR defined in Subsection \ref{image_wise_evaluation}.

    Another class-specific uncertainty performance measure proposed by \cite{mehta2020uncertainty}, has been developed for the uncertainty task of the BraTS challenge \citep{menze2014multimodal}. This performance measure was designed to reward voxels with high confidence and correct classification, reward voxels with low confidence and wrong classification and penalize the uncertainty maps that have a high proportion of under-confident correct classifications.

        The principle of this performance measure is to filter out voxels and remove them from the evaluation based on their uncertainty value. For each filtering threshold $\tau_f$, the voxels with uncertainty values in the uncertainty map $u$ above the threshold $\tau_f$ are removed from the prediction. Then, one can derive the ratio of filtered true positives ($\text{tpr}(\tau_f)$), the ratio of filtered true negatives ($\text{tnr}(\tau_f)$) and the filtered Dice score ($\text{Dice}(\tau_f)$). The BraTS uncertainty performance measure (BRATS-UNC) integrates these three measurements over the range of values of the uncertainty map, as follows:

        \begin{align}
            \text{BRATS-UNC} = \frac{1}{3} \int_{\text{min}(u)}^{\text{max}(u)} \text{Dice}(\tau_f) + (1 - \text{tnr}(\tau_f)) + (1 - \text{tpr}(\tau_f)) \mathrm{d\tau_f}.
            \label{E:bratsunc}
        \end{align}

Contrary to the AUC-PR performance measure, the BRATS-UNC performance measure take into account the calibration of the uncertainty map as $\text{tnr}$, $\text{tpr}$ and $\text{Dice}$ are integrated over the values of the uncertainty maps. Note that the BRATS-UNC is slightly altered compared to its original formulation by \cite{mehta2020uncertainty} to generalise to the case of non-normalised uncertainty maps.

\subsection{Statistical significance}\label{statisticalsignificance}

One can estimate the posterior distribution $p_{A > B}$ of the proportion of models where a given performance measure (either class-specific AUC-PR, combined AUC-PR or BRATS-UNC) has a higher average value over the test set for uncertainty map A than for uncertainty map B. In a Bayesian fashion, we choose a non-informative prior distribution of $p_{A > B} \sim \text{Beta}(1, 1)$ which corresponds to a uniform distribution. Over the $N$ models under study, we observe $k_{A > B}$ models that have a higher average value over the test set of the studied performance measure for uncertainty map A than for uncertainty map B. This observation of $k_{A > B}$ can be used, in the Bayes formula, to obtain the posterior distribution of the parameter $p_{A > B} \sim \text{Beta}(1 + k_{A > B}, 1 + N - k_{A > B})$. From this Bayesian analysis, one can derive $I_{95\%}$, the 95\% equally tailed credible interval of the parameter $p_{A > B}$, \citep{makowski2019bayestestr, mcelreath2020statistical}.

A modified version of this analysis can be performed patient-wise where $p_{A > B}$ is then the proportion of patients where a given performance measure has a higher average value over the models for uncertainty map A than for uncertainty map B. In this case, $N$ is the number of patients and $k_{A > B}$ is the number of patient with a higher average value over the models for uncertainty map A than for uncertainty map B.

\section{Experiments}

\subsection{Dataset}

We used  carotid artery MR images acquired within the multi-center, multi-scanner, multi-sequence PARISK study \citep{truijman2014plaque}, a large prospective study to improve risk stratification in patients with mild to moderate carotid artery stenosis ($<70\%$, the degree of stenosis being the ratio of the diameter of the lumen and the diameter of the vessel wall). The studied population is predominantly Caucasian (97\%). The age and gender distributions of the studied population available in Appendix F in Figure \ref{fig:dataset_statistics} show that the mean age is 69 years old (age range: [39-89], standard deviation: 9 years) and shows that a majority of the studied population is male (66\%). The standardized MR acquisition protocol is described in Appendix F in Table \ref{tab:mriconfig}. We used the images of all enrolled subjects (n=145) at three of the four study centers as these centers have used the same protocol: Amsterdam Medical Center (AMC), the Maastricht University Medical Center (MUMC), and the University Medical Center of Utrecht (UMCU), all in the Netherlands. The dataset was split with 69 patients in the training set (all from MUMC), 24 patients in the validation set (all from MUMC) and 52 patients in the test set (15 from MUMC, 24 from UMCU and 13 from AMC). Each center performed the MR imaging under 3.0-Tesla with an eight-channel phased-array coil (Shanghai Chenguang Medical Technologies Co., Shanghai). UMCU and MUMC acquired the imaging data of all the patients with an Achieva TX scanner (Phillips Healthcare, Best, Netherlands), AMC center acquired 11 of its patients with an Ingenia scanner (Phillips Healthcare, Best, Netherlands) and 2 with an Intera scanner (Phillips Healthcare, Best, Netherlands).

MR sequences were semi-automatically, first affinely and then elastically registered to the T1w pre-contrast sequence. The vessel lumen and vessel wall were annotated manually slice-wise, by trained observers with 3 years of experience, in the T1w pre-contrast sequence. Registration and annotation were achieved with VesselMass software\footnote{\url{https://medisimaging.com/apps/vesselmass-re/}}. The image intensities were linearly scaled per image such that the $5^{th}$ \% was set to 0 and the $95^{th}$ \% was set to 1. The networks were trained and tested on a region of interest of 128x128x16 voxels covering the common and internal carotid arteries. This region of interest of 128x128x16 voxels correspond to the bounding box centered on the center of mass of the annotations. The observer annotated the common and internal carotid arteries (either left or right) where the stenosis symptoms occured.

\subsection{Network implementation}\label{network}

The networks used for our experiments are based on a 3D U-net architecture as shown in Appendix G in Figure \ref{fig:network} \citep{ronneberger2015u}. Because of the low resolution of our problem in the z-axis compared to the resolution on the x-and y-axis, we  applied 2D max-pooling and 2D up-sampling slice-wise instead of their usual 3D alternatives. We trained the model using Adadelta optimizer \citep{zeiler2012adadelta} for 600 epochs with training batches of size 1. The network was optimized with the Dice loss \citep{milletari2016v}. As data augmentation, on the fly random flips along the x axis were used. The networks were implemented in Python using Pytorch \citep{paszke2019pytorch} and run on a NVIDIA GeForce 2080 RTX GPU.

\subsection{Studied parameters}

We varied three parameters in our network : the number of images in the training sample to analyse the robustness of the performance measure to networks with different level of segmentation performances, the dropout rate, and the dropout type to test different variations of MC dropout. Eight values of number of images in the training set were used : 3, 5, 9, 15, 25, 30, 40 and 69 images. Also, nine dropout rates were used at train and test time: 0.1, 0.2, 0.3, 0.4, 0.5, 0.6, 0.7, 0.8 and 0.9. Finally the two types of dropout (Bernoulli dropout and Gaussian multiplicative noise) described in subsection \ref{mcdropout} were considered. For every combination of those three parameters, we trained a network following the procedure detailed in Subsection \ref{network}. The total number of trained models used in the evaluation is 144. We discretized the integrals of Equation \ref{E:Dh}, \ref{E:Sb} and \ref{E:bratsunc} in $n_{bins} = 100$ bins using a left Riemann sum approximation and we sampled $T=50$ times using MC dropout method.

\begin{figure}[t]
    \centering
    \foreach \outputclass in {Background, Vessel-Wall, Lumen}{
		\begin{subfigure}[b]{0.3\textwidth}
		    \includegraphics[width=\textwidth]{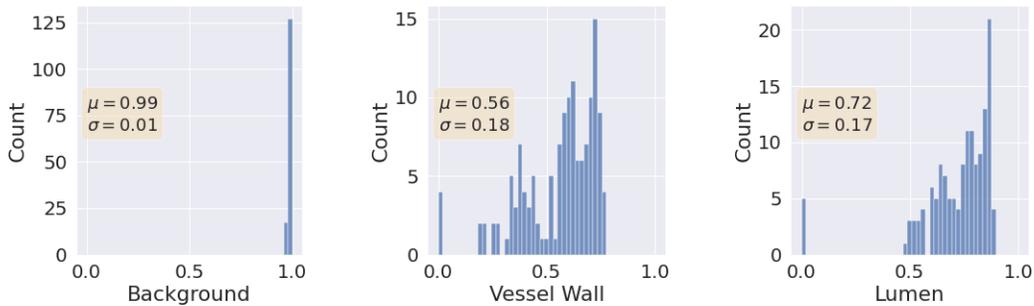}
		\end{subfigure}
    }
    \caption{Distribution over the 144 models of the Dice coefficient averaged over the test set, for each of the three classes.}
    \label{fig:result_dice}
\end{figure}

\section{Results}

\begin{figure}[t]
    \centering
    \foreach \file in {training_set_size, dropout_rate, dropout_type, legend}{
		\begin{subfigure}[b]{0.49\textwidth}
		    \includegraphics[width=\textwidth]{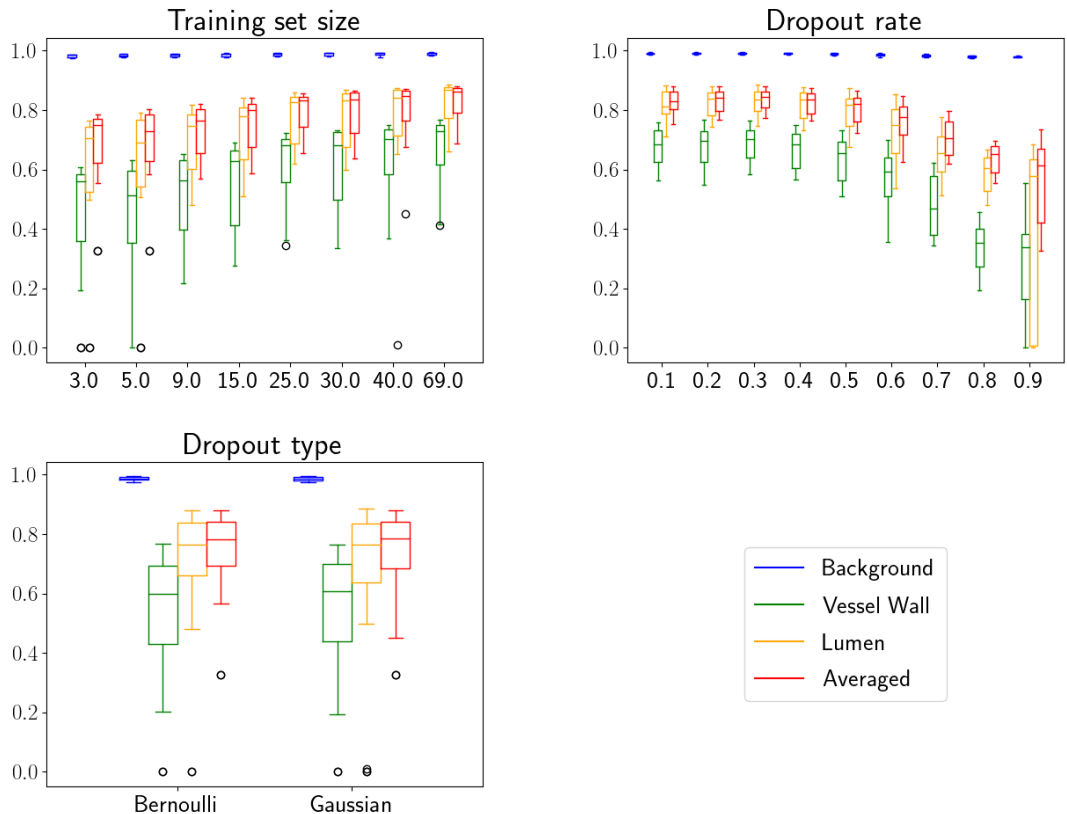}
		\end{subfigure}
    }
    \caption{Distributions of the Dice scores averaged over the test set aggregated per parameter (training set size, dropout type, dropout rate). The presented Dice score are either computed per class (background, vessel wall, lumen) or averaged over the classes (averaged)}
    \label{fig:dice_per_parameter}
\end{figure}

In this section, we first present the results of the different models in the segmentation task. This will be followed by the qualitative and quantitative results of the combined uncertainty maps. Finally, qualitative and quantitative results of class-specific uncertainty maps will be shown for both performance metrics under study (class-specific AUC-PR and BRATS-UNC).

\subsection{Segmentation}

One can find the distribution of the Dice segmentation overlap per class averaged over the test set in Figure \ref{fig:result_dice}. The highest averaged Dice over classes was observed with a model trained with Gaussian multiplicative noise and a dropout rate of 0.3 on the whole training set (69 samples). This method achieved Dice scores of 0.994 on the background, 0.764 on the vessel wall and 0.885 on the lumen. On the other end of the spectrum, the 3 models that did not converge are the models with a dropout rate of 0.9, a training set size 3 for both Gaussian multiplicative noise and Bernoulli dropout and the model with Gaussian multiplicative noise, a dropout rate of 0.9 and a training set size of 5. Examples of segmentation performances for different models are displayed in the prediction row of Figure \ref{fig:multiclassvizualisation}, \ref{fig:backgroundvizualisation}, \ref{fig:vesselwallvizualisation} and \ref{fig:lumenvizualisation}. The effect of the different hyperparameters is shown in Figure \ref{fig:dice_per_parameter}. As expected, better performance is observed with an increasing dataset size while a high value of dropout harms the prediction. It also shows that very similar results are observed for both Gaussian multiplicative noise and Bernoulli dropout.

\begin{figure}[t]
        \includegraphics[width=\textwidth]{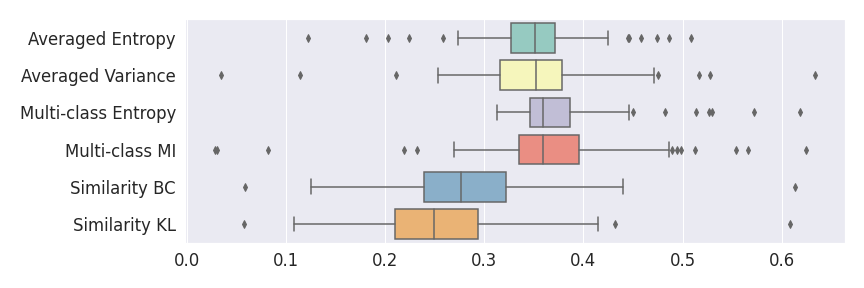}
        \caption{Distribution over the 144 models of the combined AUC-PR averaged over test set for each uncertainty map. The whiskers represent the 5\% and 95\% interval.}
        \label{fig:distributionimagewiseaucpr}
\end{figure}

\begin{table}[t]
    \caption{95\% credible interval of the posterior distribution of $p_{A > B}$ for the combined AUC-PR performance measure and the combined uncertainty maps pairs A and B. The rows correspond to the uncertainty maps A and the columns to the uncertainty map B. The statistically significant differences are reported in bold ($0.5 \notin I_{95\%}$). $p_{A > B} > 0.5$ means that uncertainty map A is better than uncertainty map B and the contrary if $p_{A > B} < 0.5$ (Av = Averaged, Mu = Multi-class, Sim = Similarity)}
    \label{tab:multiclasssignificance}
    \centering
    \begin{tabular}{p{0.14\textwidth}|p{0.14\textwidth}|p{0.14\textwidth}|p{0.14\textwidth}|p{0.14\textwidth}|p{0.14\textwidth}}
\hline
& Av Variance &     Mu Entropy &          Mu MI &           Sim BC &           Sim KL \\
\hline
Av Entropy    &    $[0.43, 0.59]$ &   $\textbf{[0.20, 0.35]}$ &  $\textbf{[0.29, 0.45]}$ &  $\textbf{[0.89, 0.97]}$ &   $\textbf{[0.90, 0.98]}$ \\
Av Variance   &                   &  $\textbf{[0.22, 0.37]}$ &   $\textbf{[0.30, 0.46]}$ &  $\textbf{[0.93, 0.99]}$ &  $\textbf{[0.91, 0.98]}$ \\
Mu Entropy &                   &                         &          $[0.49, 0.65]$ &  $\textbf{[0.93, 0.99]}$ &   $\textbf{[0.95, 1.00]}$ \\
Mu MI      &                   &                         &                         &  $\textbf{[0.89, 0.97]}$ &  $\textbf{[0.91, 0.98]}$ \\
Sim BC       &                   &                         &                         &                         &  $\textbf{[0.64, 0.79]}$ \\
\hline
\end{tabular}

\end{table}

\subsection{Combined uncertainty maps evaluation} \label{image_wise_results}

Figure \ref{fig:distributionimagewiseaucpr} describes the distribution of the combined AUC-PR averaged over the test set. The $95\%$ equally tailed credible interval of the posterior distribution of $p_{A > B}$, the proportion of models that have a higher combined AUC-PR average value over the test set for uncertainty map A than for uncertainty map B, is reported in Table \ref{tab:multiclasssignificance}. A value of $p_{A > B}$ superior to $0.5$ indicates that uncertainty map A performs better than uncertainty map B and similarly a value lower to $0.5$ indicates the opposite. A pair-wise comparison indicates a statistically significant difference if the $95\%$ equally tailed credible interval does not contain the expected random behaviour ($p_{A > B} = 0.5$). Twenty-six out of the thirty pair-wise comparisons are statistically significant ($0,5 \notin I_{95\%}$). The statistically significant best results are observed with the multi-class entropy and the multi-class mutual information. The statistically significant worst result is observed with the similarity KL. The different posterior distributions of $p_{A > B}$ are reported in Figure\ref{fig:statsignmulticlass} in Appendix C.

Figure \ref{fig:multiclassvizualisation} provides a qualitative visualization of the different uncertainty maps for different models and patients. All the uncertainty maps shows higher value of uncertainty at the inner and outer boundaries of the vessel wall which is to be expected as it corresponds to difficult areas to segment. Also for each uncertainty map the level of uncertainty decreases when performance increases. The similarity KL tend to mark as equally uncertain large regions of the image which explains its bad performance.

Figure \ref{fig:combined_pr_curve} in Appendix G presents the combined precision recall curves averaged over the patients of the test set for the best model (model with gaussian multiplicative noise, a dropout rate of 0.3 trained on the whole training set). This plots highlight the poor performance of both uncertainty maps using the Similarity aggregation method.

To assess generalisability of our results, a similar statistical analysis was conducted individually for the different centers individually (AMC, UMCU and MUMC) and for the different types of dropout (Bernouilli dropout or Gaussian Multiplicative Noise). Figure \ref{fig:center_correlation} in Appendix F presents the correlation plots of the mode of the posterior distribution of $p_{A > B}$ obtained on the whole test set and on the studied subset (AMC, UMCU, MUMC, Bernouilli dropout and Gaussian Multiplicative Noise). The statistical analysis generalises well to a subset if the points of the correlation plot are close to the line $y=x$.

Table \ref{tab:top10multiclasssignificance} in Appendix H presents a modified version of statistical analysis. In this analysis instead of all the models only the 10 best models are considered (higher averaged dice score on the test set). Due to the small amount of experiments considered the combined AUC-PR is not aggregated experiment-wise but patient-wise. $p_{A, B}$ correspond then to the proportion of patients with a higher combined AUC-PR averaged over the 10 best models for uncertainty map A than for uncertainty map B. Similar conclusion can be drawn regarding the best and worst combined uncertainty maps than in the main statistical analysis.

\subsection{Class-specific uncertainty maps comparison}

\begin{figure}[t]
        \includegraphics[width=\textwidth]{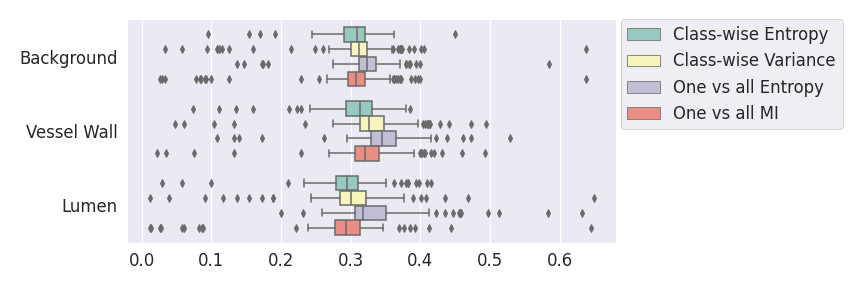}
        \caption{Distribution over the 144 models of the class-specific AUC-PR averaged over the test set for each class. The whiskers represent the 5\% to 95\% interval.}
        \label{fig:distributionclasswiseaucpr}
\end{figure}

\begin{figure}[t]
        \includegraphics[width=\textwidth]{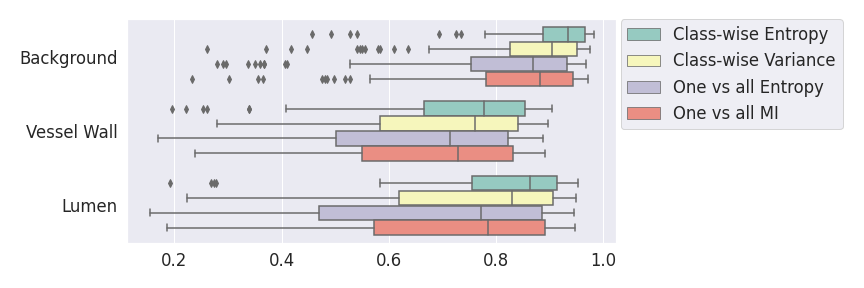}
        \caption{Distribution over the 144 models of the BRATS-UNC performance measure averaged over the test set for each class. The whiskers represent the 5\% to 95\% interval.}
        \label{fig:distributionbrats}
\end{figure}

\begin{table}[t]
\caption{95\% credible interval of the posterior distribution of $p_{A > B}$ for the class-specific AUC-PR performance measure and the class-specific uncertainty maps pairs A and B. The rows correspond to the uncertainty maps A and the columns to the uncertainty map B. The statistically significant results are reported in bold ($0.5 \notin I_{95\%}$). $p_{A, B} > 0.5$ means that uncertainty map A is better than uncertainty map B and the contrary if $p_{A > B} < 0.5$. (CW = Class-wise, 1vA = One versus all)}
\label{tab:classwiseaucprsignificance}
\centering
\subcaption*{Background}
\begin{tabular}{p{0.167\textwidth}|p{0.167\textwidth}|p{0.167\textwidth}|p{0.167\textwidth}}
\hline
{} &     CW variance &      1vA Entropy &           1vA MI \\
\hline
CW Entropy  &  $\textbf{[0.31, 0.46]}$ &   $\textbf{[0.09, 0.20]}$ &          $[0.39, 0.55]$ \\
CW variance &                         &  $\textbf{[0.07, 0.17]}$ &   $\textbf{[0.90, 0.98]}$ \\
1vA Entropy  &                         &                         &  $\textbf{[0.87, 0.96]}$ \\
\hline
\end{tabular}

\subcaption*{Vessel Wall}
\begin{tabular}{p{0.167\textwidth}|p{0.167\textwidth}|p{0.167\textwidth}|p{0.167\textwidth}}
\hline
{} &     CW variance &      1vA Entropy &           1vA MI \\
\hline
CW Entropy  &  $\textbf{[0.26, 0.41]}$ &  $\textbf{[0.03, 0.11]}$ &          $[0.36, 0.52]$ \\
CW variance &                         &  $\textbf{[0.03, 0.11]}$ &  $\textbf{[0.94, 0.99]}$ \\
1vA Entropy  &                         &                         &  $\textbf{[0.92, 0.98]}$ \\
\hline
\end{tabular}

\subcaption*{Lumen}
\begin{tabular}{p{0.167\textwidth}|p{0.167\textwidth}|p{0.167\textwidth}|p{0.167\textwidth}}
\hline
{} &     CW Variance &      1vA Entropy &          1vA MI \\
\hline
CW Entropy  &  $\textbf{[0.33, 0.48]}$ &  $\textbf{[0.06, 0.16]}$ &         $[0.45, 0.61]$ \\
CW Variance &                         &  $\textbf{[0.01, 0.06]}$ &  $\textbf{[0.95, 1.00]}$ \\
1vA Entropy  &                         &                         &  $\textbf{[0.95, 1.00]}$ \\
\hline
\end{tabular}

\end{table}

\begin{table}[t]
\caption{95\% credible interval of the posterior distribution of $p_{A > B}$ for the BRATS-UNC performance measure per class and the class-specific uncertainty maps pairs A and B.The rows correspond to the uncertainty maps A and the columns to the uncertainty map B. The statistically significant results are reported in bold ($0.5 \notin I_{95\%}$). $p_{A > B} > 0.5$ means that uncertainty map A is better than uncertainty map B and the contrary if $p_{A > B} < 0.5$ (CW = Class-wise, 1vA = One versus all)}
\label{tab:classwisesignificance}
\centering
\subcaption*{Background}
\begin{tabular}{p{0.167\textwidth}|p{0.167\textwidth}|p{0.167\textwidth}|p{0.167\textwidth}}
\hline
{} &     CW Variance &     1vA Entropy &          1vA MI \\
\hline
CW Entropy  &  $\textbf{[0.89, 0.97]}$ &  $\textbf{[0.96, 1.00]}$ &  $\textbf{[0.96, 1.00]}$ \\
CW Variance &                         &  $\textbf{[0.96, 1.00]}$ &  $\textbf{[0.97, 1.00]}$ \\
1vA Entropy  &                         &                        &  $\textbf{[0.00, 0.04]}$ \\
\hline
\end{tabular}

\subcaption*{Vessel Wall}
\begin{tabular}{p{0.167\textwidth}|p{0.167\textwidth}|p{0.167\textwidth}|p{0.167\textwidth}}
\hline
{} &    CW Variance &     1vA Entropy &           1vA MI \\
\hline
CW Entropy  &  $\textbf{[0.79, 0.90]}$ &  $\textbf{[0.95, 1.00]}$ &  $\textbf{[0.89, 0.97]}$ \\
CW Variance &                        &  $\textbf{[0.97, 1.00]}$ &   $\textbf{[0.96, 1.00]}$ \\
1vA Entropy  &                        &                        &   $\textbf{[0.00, 0.03]}$ \\
\hline
\end{tabular}

\subcaption*{Lumen}
\begin{tabular}{p{0.167\textwidth}|p{0.167\textwidth}|p{0.167\textwidth}|p{0.167\textwidth}}
\hline
{} &     CW Variance &     1vA Entropy &           1vA MI \\
\hline
CW Entropy  &  $\textbf{[0.84, 0.94]}$ &  $\textbf{[0.96, 1.00]}$ &  $\textbf{[0.94, 0.99]}$ \\
CW Variance &                         &  $\textbf{[0.97, 1.00]}$ &  $\textbf{[0.94, 0.99]}$ \\
1vA Entropy  &                         &                        &   $\textbf{[0.00, 0.04]}$ \\
\hline
\end{tabular}

\end{table}

In a class-specific case, two performance measures are under study, the class-specific AUC-PR and the BRATS-UNC. Figures \ref{fig:distributionclasswiseaucpr} and \ref{fig:distributionbrats} describe the distribution of the average value over the test set for the different models of the class-specific AUC-PR performance measure and of the BRATS-UNC performance measure respectively. The $95\%$ equally tailed credible interval of $p_{A > B}$, the proportion of models that have a higher value of AUC-PR for uncertainty map A than uncertainty map B, is reported in Table \ref{tab:classwiseaucprsignificance}; a similar table is provided for the BRATS-UNC performance measure in Table \ref{tab:classwisesignificance}. Ten out of the twelve pair-wise comparisons of uncertainty maps for the different classes showed significant differences for the class-specific AUC-PR and all of the pair-wise comparisons showed significant differences for the BRATS-UNC performance measure. For the three classes under-study, the best results were observed for the class-specific AUC-PR with the one versus all entropy and the worst with the class-wise entropy. The best results were observed for the BRATS-UNC performance measure with the class-wise entropy and the worst with the one versus all entropy.

The posterior distribution $p_{A > B}$ for the class-specific AUC-PR performance measure can be respectively found for the background, the vessel wall and the lumen in Figure \ref{fig:statsignaucprbackground}, \ref{fig:statsignaucprvesselwall} and \ref{fig:statsignaucprlumen} of Appendix D. Similarly, the posterior distribution $p_{A > B}$ for the BRATS-UNC performance measure can be respectively found for the background, the vessel wall and the lumen in Figure \ref{fig:statsignbackground}, \ref{fig:statsignvesselwall} and \ref{fig:statsignlumen} of Appendix E. 

\begin{figure}[H]
    \centering
    \includegraphics[width=.95\textwidth]{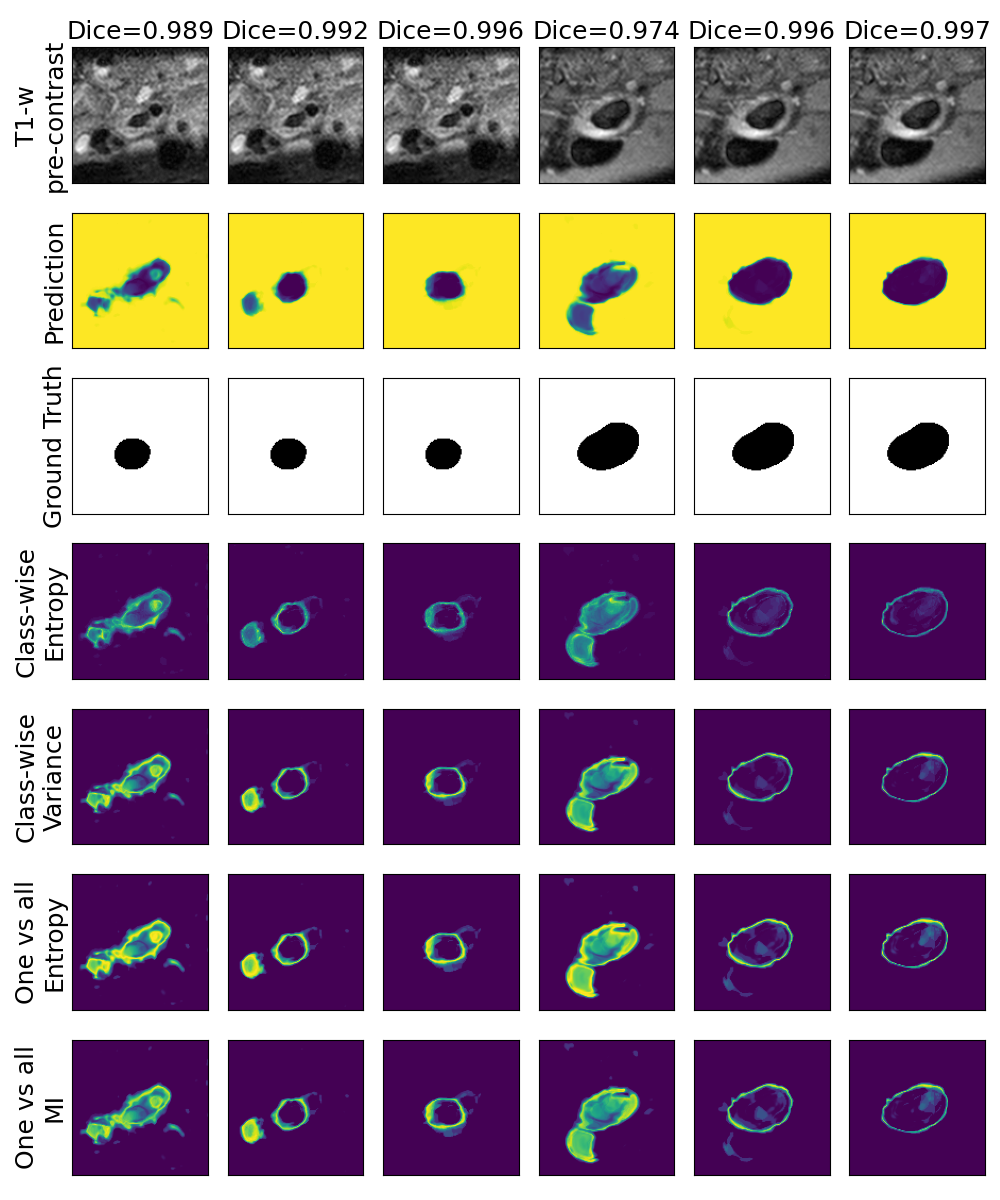}
    \caption{Example of the different uncertainty measures corresponding to the background class. From up to down, rows represent: the T1w pre-contrast MR image, the background prediction, the background ground truth and the different uncertainty maps (Class-wise Entropy, Class-wise Variance, One versus all Entropy, One vs all MI). The columns correspond to predictions with 2 different patients and with 3 different networks. The indicated Dice is the Dice of the background class}
    \label{fig:backgroundvizualisation}
\end{figure}

\begin{figure}[H]
     \centering
     \includegraphics[width=.95\textwidth]{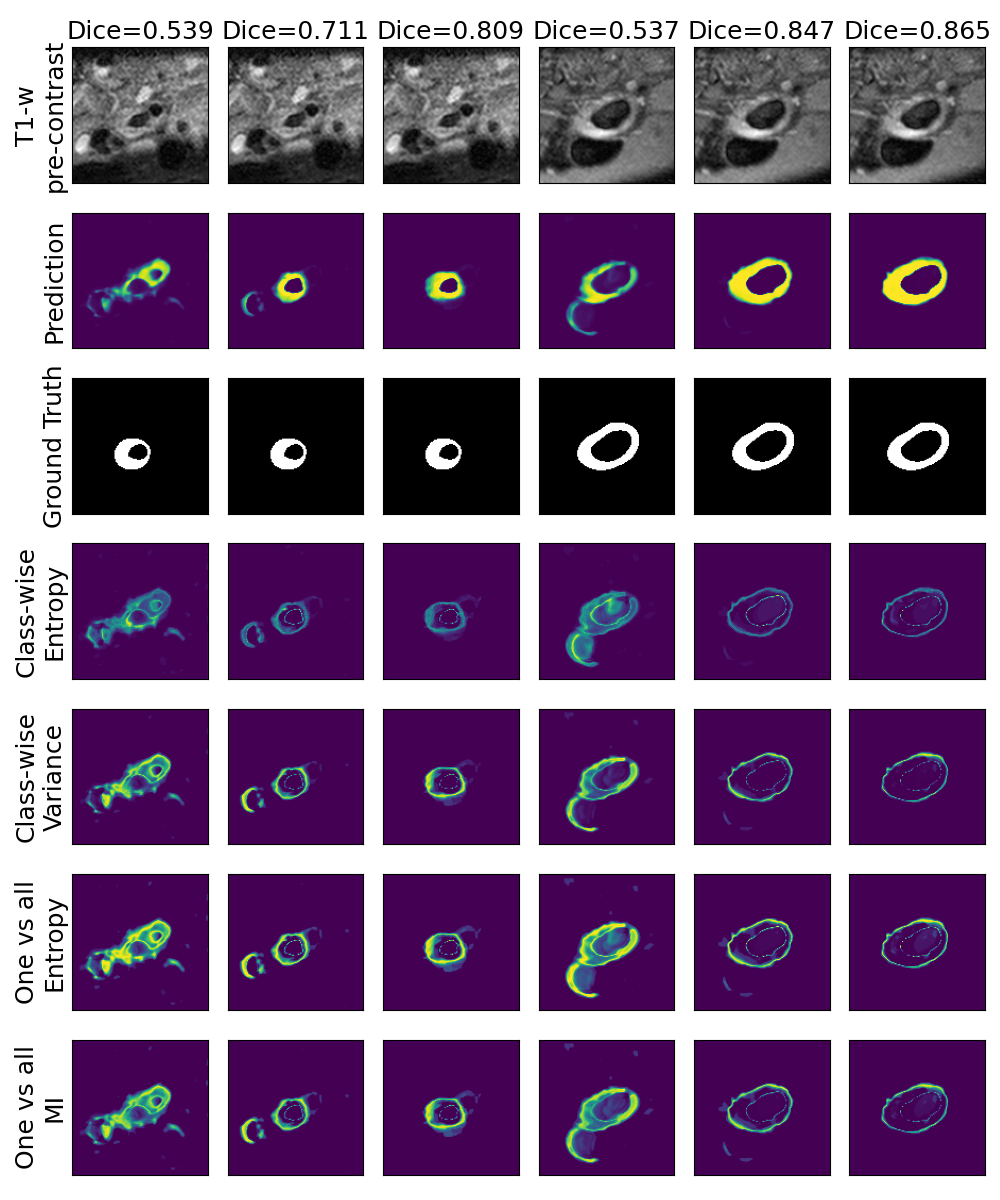}
     \caption{Example of the different uncertainty measures corresponding to the vessel wall class. From up to down, rows represent: the T1w pre-contrast MR image, the vessel wall prediction, the vessel wall ground truth and the different uncertainty maps (Class-wise Entropy, Class-wise Variance, One versus all Entropy, One vs all MI). The columns correspond to predictions with 2 different patients and with 3 different networks. The indicated Dice is the Dice of the vessel wall class}
     \label{fig:vesselwallvizualisation}
\end{figure}

\begin{figure}[H]
     \centering
     \includegraphics[width=.95\textwidth]{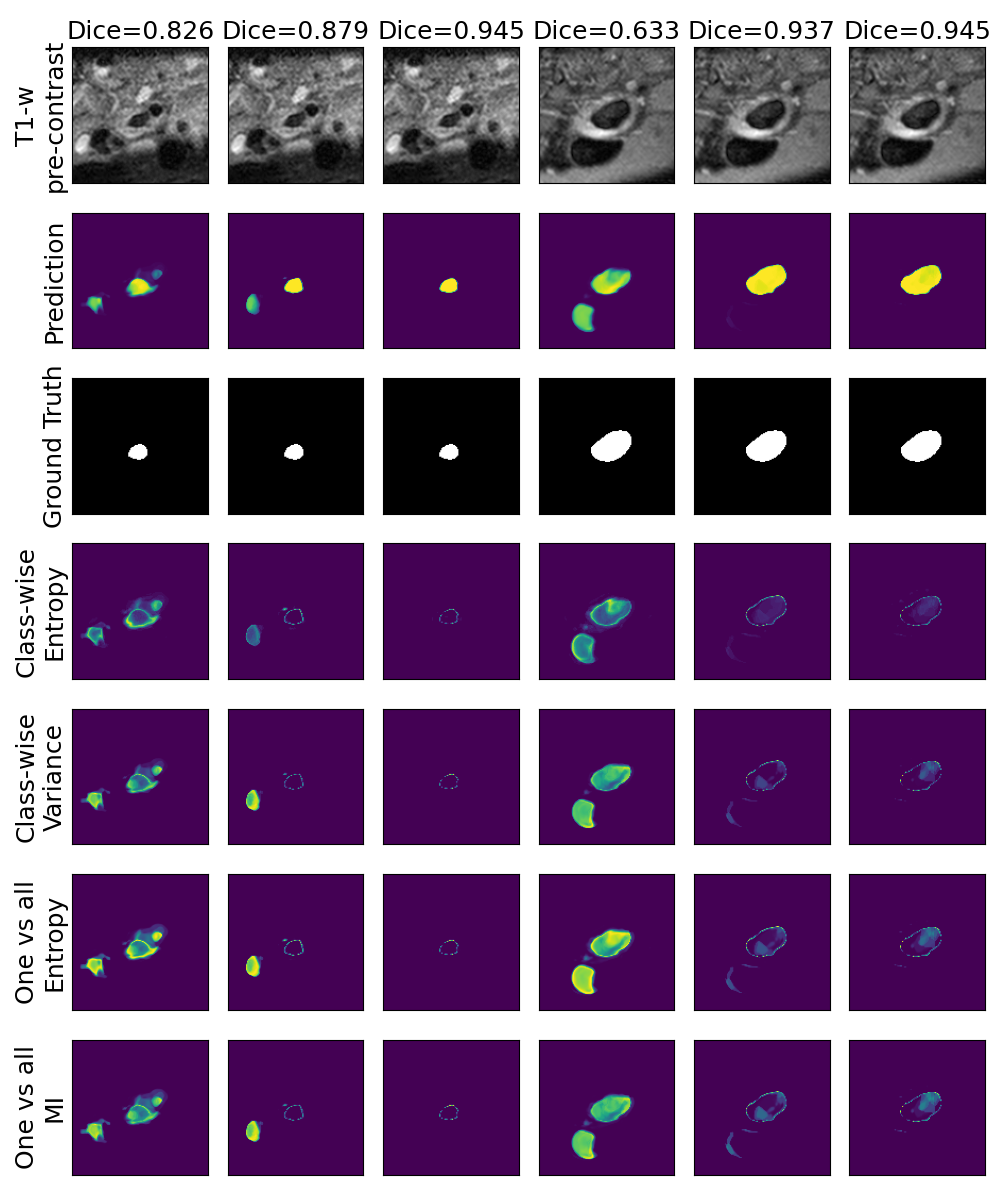}
     \caption{Example of the different uncertainty measures corresponding to the lumen class. From up to down, rows represent: the T1w pre-contrast MR image, the lumen prediction, the lumen ground truth and the different uncertainty maps (Class-wise Entropy, Class-wise Variance, One versus all Entropy, One vs all MI). The columns correspond to predictions with 2 different patients and with 3 different networks. The indicated Dice is the Dice of the lumen class}
     \label{fig:lumenvizualisation}
\end{figure}

 A visualization of the different uncertainty maps for different models and patients is available in Figure \ref{fig:backgroundvizualisation}for the background class, in Figure \ref{fig:vesselwallvizualisation} for the vessel wall class and in Figure \ref{fig:lumenvizualisation} for the lumen class. As for the combined uncertainty maps, the level of uncertainty decreases when the level of performance increases. The most uncertain areas are the boundaries of the considered class (either background, vessel wall and lumen). This is coherent as it corresponds to difficult areas to segment. Those qualitative results also show that for all classes the class-wise entropy have the lowest level of uncertainty as it is the most hypo-intense uncertainty map and the one-vs-all entropy have the highest level of uncertainty as it is the most hyper-intense uncertainty map.

Figure \ref{fig:curves} in Appendix G presents the class-specific precision recall curves and the component integrated in the BRATS-UNC performance measure averaged over the patients of the test set for the best model (model with gaussian multiplicative noise, a dropout rate of 0.3 trained on the whole training set). This figure shows that the tpr curve dominate the integral for minority classes (vessel wall and lumen) and tnr dominate the integral for the background class. The curvature of the vessel-wall tpr and the background tnr curves increases for uncertainty maps with a lower level of uncertainty such as in the class-wise entropy. This confirms that contrary to class-specific AUC-PR, BRATS-UNC assesses the calibration of the uncertainty map.

To assess generalisability of our results, a similar statistical analysis was conducted individually for the different centers individually (AMC, UMCU and MUMC) and for the different types of dropout (Bernouilli dropout or Gaussian Multiplicative Noise). Figure \ref{fig:center_correlation} in Appendix F presents the correlation plots of the mode of the posterior distribution of $p_{A > B}$ obtained on the whole test set and on the studied subset (AMC, UMCU, MUMC, Bernouilli dropout and Gaussian Multiplicative Noise). The statistical analysis generalises well to a subset if the points of the correlation plot are close to the line $y=x$.

Tables \ref{tab:top10classwiseaucprsignificance} and \ref{tab:top10classwisesignificance} in Appendix H presents a modified version of statistical analysis for class-specific AUC-PR and BRATS-UNC respectively. In this analysis instead of all the models only the 10 best models are considered (higher averaged dice score on the test set). Due to the small amount of experiments considered the studied performance measure (either class-specific AUC-PR or BRATS-UNC) are not aggregated experiment-wise but patient-wise. $p_{A, B}$ correspond then to the proportion of patients with a higher value of the studied performance measure averaged over the the 10 best models for uncertainty map A than for uncertainty map B. For both performance measures and all classes, similar conclusion can be drawn regarding the best uncertainty maps than in the statistical analyses presented in Tables \ref{tab:classwiseaucprsignificance} and \ref{tab:classwisesignificance}.

\begin{figure}[H]
    \centering
    \includegraphics[width=.78\textwidth]{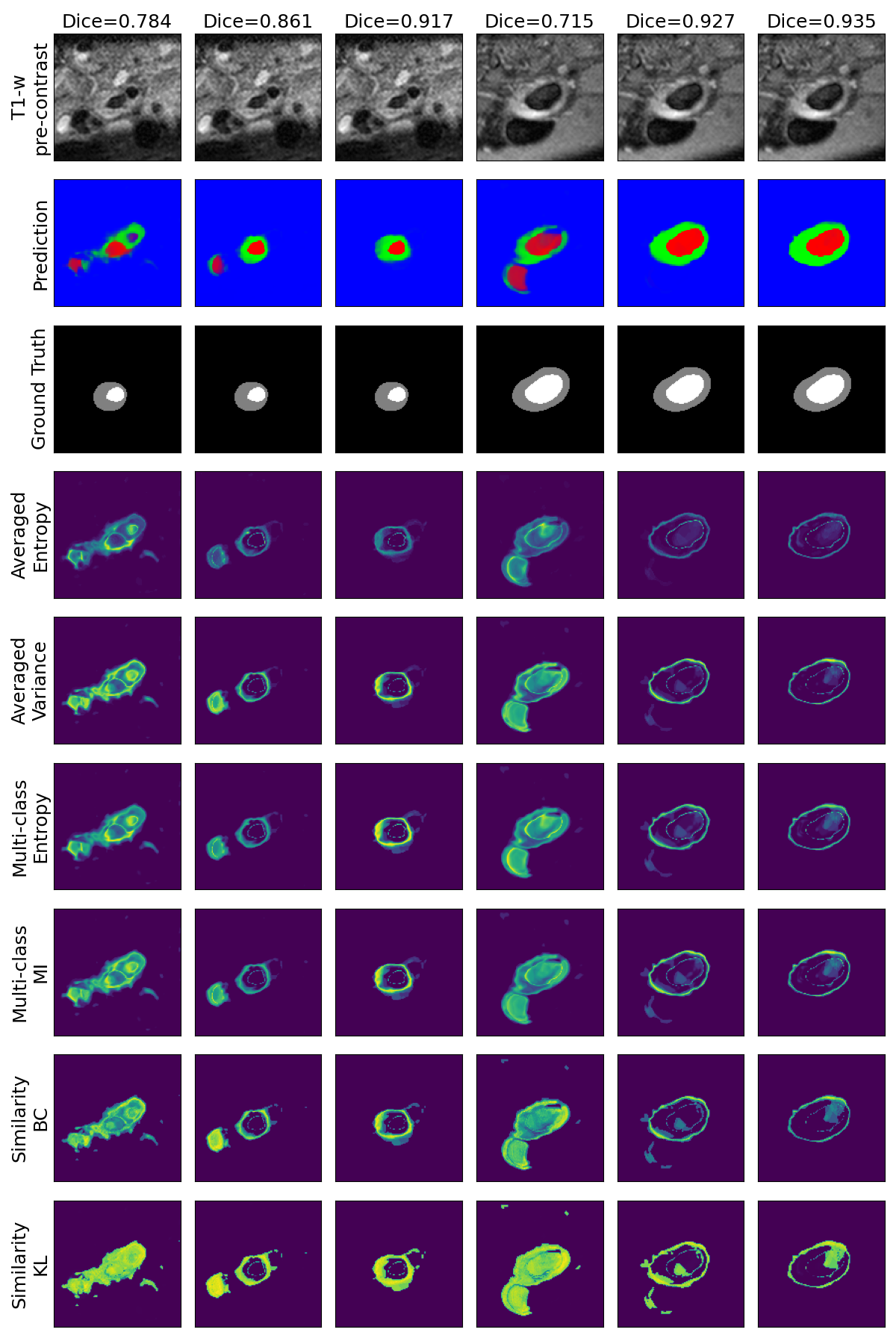}
    \caption{Example of the different uncertainty measures in a combined case. From top to bottom, rows represent: the T1w pre-contrast MR image, the multi-class prediction (blue = background, green = vessel wall, red = lumen, the level of brightness corresponds to the probability of the predicted class), the ground truth and the different uncertainty maps (Averaged Entropy, Averaged Variance, Multi-class Entropy, Mutual MI, Similarity BC and Similarity KL). Columns 1--3 and 4--6  correspond to predictions with 2 different patients and columns 1 and 4, 2 and 5, 3 and 6 correspond to 3 different networks. The indicated Dice is the averaged Dice coefficient over classes}
    \label{fig:multiclassvizualisation}
\end{figure}

\section{Discussion}



In a combined scenario, considering the relationship between the order of the voxels sorted according to their epistemic uncertainty values and the misclassification of the prediction (using combined AUC-PR performance measure), the multi-class entropy and multi-class mutual information statistically out-perform the averaged variance and the averaged entropy which in turn statistically out-perform the similarity BC which out-performs the similarity KL. This analysis highlights the good performances of the multi-class aggregation method and the averaged aggregation method compared with the similarity aggregation method which is coherent with their extensive use in the literature \citep{kendall2015bayesian, seebock2019exploiting, mobiny2019risk, mobiny2019dropconnect}. The superiority of the multi-class aggregation method over the class-wise aggregation method seems also coherent as the averaged aggregation method consists of a combination of class-specific uncertainty maps where the multi-class aggregation method is designed for multi-class problems.


With similar considerations for class-specific uncertainty maps, the one versus all entropy statistically out-performs the class-wise variance which in turn statistically out-performs the other uncertainty maps under study. We can then conclude that with these considerations, the entropy uncertainty measure out-performs the other uncertainty measures with a multi-class aggregation method in a combined scenario and with a one versus all aggregation method in a class-specific scenario.

In our class-specific analysis, we compared our performance measure (class-specific AUC-PR) to the performance measure used in the BraTS challenge (BRATS-UNC) which assesses the calibration of an uncertainty map. This comparison highlights the four following points.  Firstly, the ranking of the different uncertainty maps for a given performance measure (either BRATS-UNC or class-specific AUC-PR) does not depend on the considered class as, for the three classes, the uncertainty maps that gave the best results are the same and the statistical comparisons of uncertainty maps are similar.  Secondly, the range of values of the class-specific AUC-PR is similar for different classes whereas it varies wildly between classes for the BRATS-UNC performance measure. This is because the three components integrated by the BRATS-UNC depends on the class imbalance. Thirdly, the ranking of uncertainty maps is different with the class-specific AUC-PR and the BRATS-UNC performance measures. The best results are observed with the one versus all entropy for the class-specific AUC-PR performance measure and with the class-specific entropy for the BRATS-UNC performance measure. A plausible explanation is that class-specific AUC-PR only considers the correct order of voxels sorted according to their epistemic uncertainty values where BRATS-UNC also measures the calibration of the uncertainty map. In that case, the BRATS-UNC penalizes the hyper-intensity of the class-wise entropy and rewards the hypo-intensity of the one versus all entropy in the region correctly segmented. Forthly, similarly to \cite{nair2020exploring}, our qualitative results shows that class-wise variance and the one versus all mutual information appears similar in term of calibration. For uncertainty maps with similar calibrations, it is coherent that the BRATS-UNC focusses on the order of the voxels sorted according to their epistemic uncertainty values and therefore gives the same results as the class-specific AUC-PR.

A limitation of our analysis is the computation of the entropy defined in Equations \ref{E:Dh} and of the Bhattacharya coefficient defined in Equation \ref{E:Sb}. This computation introduce an extra hyper-parameter, the discretization step of the integral. A too high or too low value of this discretization step will lead to a poor approximation of the uncertainty measure. The value of this hyper-parameter was set to the value found in the literature in  \cite{de2019quantifying} but further tuning might lead to better results. However, the value set in this article seems reasonable as the best result in the class-specific case with BRATS-UNC performance measure is observed with the class-wise entropy that is subject to that discretization.

In this article, we proposed a systematic approach to characterize epistemic uncertainty maps that can be used in a multi-class segmentation context. We also proposed a methodology to analyse the quality of multi-class epistemic uncertainty maps. Both the systematic approach and the methodology are not data specific or task specific. Therefore, to reproduce our work on other datasets and tasks, we made a python package available on pypi platform.

\section{Conclusion}

In conclusion, we proposed a systematic approach to characterize epistemic uncertainty maps. We validated our methodology over a large set of trained models to compare ten uncertainty maps: six combined uncertainty maps and four class-specific uncertainty maps. Our analysis was applied to the multi-class segmentation of the vessel-wall and the lumen of the carotid artery applied to a multi-center, multi-scanner, multi-sequence study. For our application and assessing the relationship between the order of the voxels sorted according to their epistemic uncertainty values and the misclassification, the best combined uncertainty maps observed are the multi-class entropy and the multi-class mutual information and the best class-specific uncertainty maps observed is one versus all entropy. Considering the calibration of the uncertainty map, the best results are observed with the class-wise entropy. Our systematic approach, being data and task independent, could be reproduced on any multi-class segmentation problem.

\vspace{1em}

\acks{This work was funded by Netherlands Organisation for Scientific Research (NWO) VICI project VI.C.182.042. The PARISK study was funded within the framework of CTMM,the Center for Translational Molecular Medicine, project PARISK (grant 01C-202), and supported by the Dutch Heart Foundation.}

\vspace{1em}

\ethics{The work follows appropriate ethical standards in conducting research and writing the manuscript, following all applicable laws and regulations regarding treatment of animals or human subjects. Due to the nature of this research, participants of this study did not agree for their data to be shared publicly, so supporting data is not available.}

\coi{The authors declare no conflicts of interest.}

\bibliography{lit}

\begin{thebibliography}{51}
\providecommand{\natexlab}[1]{#1}
\providecommand{\url}[1]{\texttt{#1}}
\expandafter\ifx\csname urlstyle\endcsname\relax
  \providecommand{\doi}[1]{doi: #1}\else
  \providecommand{\doi}{doi: \begingroup \urlstyle{rm}\Url}\fi

\bibitem[Abdar et~al.(2020)Abdar, Pourpanah, Hussain, Rezazadegan, Liu,
  Ghavamzadeh, Fieguth, Khosravi, Acharya, Makarenkov, and
  Nahavandi]{abdar2020review}
Moloud Abdar, Farhad Pourpanah, Sadiq Hussain, Dana Rezazadegan, Li~Liu,
  Mohammad Ghavamzadeh, Paul Fieguth, Abbas Khosravi, U~Rajendra Acharya,
  Vladimir Makarenkov, and Saeid Nahavandi.
\newblock {A Review of Uncertainty Quantification in Deep Learning: Techniques,
  Applications and Challenges}.
\newblock \emph{arXiv preprint arXiv:2011.06225}, 2020.

\bibitem[Arias~Lorza et~al.(2018)Arias~Lorza, Van~Engelen, Petersen, Van
  Der~Lugt, and de~Bruijne]{arias2018maximization}
Andres~M Arias~Lorza, Arna Van~Engelen, Jens Petersen, Aad Van Der~Lugt, and
  Marleen de~Bruijne.
\newblock {Maximization of regional probabilities using Optimal Surface Graphs:
  Application to carotid artery segmentation in MRI}.
\newblock \emph{Medical physics}, 45\penalty0 (3):\penalty0 1159--1169, 2018.

\bibitem[Blum et~al.(2019)Blum, Sarlin, Nieto, Siegwart, and
  Cadena]{blum2019fishyscapes}
Hermann Blum, Paul-Edouard Sarlin, Juan Nieto, Roland Siegwart, and Cesar
  Cadena.
\newblock {The Fishyscapes Benchmark: Measuring Blind Spots in Semantic
  Segmentation}.
\newblock \emph{arXiv e-prints}, art. arXiv:1904.03215, April 2019.

\bibitem[Blundell et~al.(2015)Blundell, Cornebise, Kavukcuoglu, and
  Wierstra]{blundell2015weight}
Charles Blundell, Julien Cornebise, Koray Kavukcuoglu, and Daan Wierstra.
\newblock {Weight Uncertainty in Neural Network}.
\newblock In Francis Bach and David Blei, editors, \emph{Proceedings of the
  32nd International Conference on Machine Learning}, volume~37 of
  \emph{Proceedings of Machine Learning Research}, pages 1613--1622, Lille,
  France, 07--09 Jul 2015. PMLR.
\newblock URL \url{http://proceedings.mlr.press/v37/blundell15.html}.

\bibitem[Camarasa et~al.(2020)Camarasa, Bos, Hendrikse, Nederkoorn, Kooi,
  van~der Lugt, and de~Bruijne]{camarasa2020quantitative}
Robin Camarasa, Daniel Bos, Jeroen Hendrikse, Paul Nederkoorn, Eline Kooi, Aad
  van~der Lugt, and Marleen de~Bruijne.
\newblock {Quantitative Comparison of Monte-Carlo Dropout Uncertainty Measures
  for Multi-class Segmentation}.
\newblock In Carole~H. Sudre, Hamid Fehri, Tal Arbel, Christian~F. Baumgartner,
  Adrian Dalca, Ryutaro Tanno, Koen Van~Leemput, William~M. Wells, Aristeidis
  Sotiras, Bartlomiej Papiez, Enzo Ferrante, and Sarah Parisot, editors,
  \emph{Uncertainty for Safe Utilization of Machine Learning in Medical
  Imaging, and Graphs in Biomedical Image Analysis}, pages 32--41, Cham, 2020.
  Springer International Publishing.
\newblock ISBN 978-3-030-60365-6.

\bibitem[Chotzoglou and Kainz(2019)]{chotzoglou2019exploring}
Elisa Chotzoglou and Bernhard Kainz.
\newblock {Exploring the Relationship Between Segmentation Uncertainty,
  Segmentation Performance and Inter-observer Variability with Probabilistic
  Networks}.
\newblock In Luping Zhou, Nicholas Heller, Yiyu Shi, Yiming Xiao, Raphael
  Sznitman, Veronika Cheplygina, Diana Mateus, Emanuele Trucco, Xiaobo~Sharon
  Hu, Danny~Ziyi Chen, Matthieu Chabanas, Hassan Rivaz, and Ingerid Reinertsen,
  editors, \emph{Large-Scale Annotation of Biomedical Data and Expert Label
  Synthesis and Hardware Aware Learning for Medical Imaging and Computer
  Assisted Intervention - International Workshops, {LABELS} 2019, {HAL-MICCAI}
  2019, and CuRIOUS 2019, Held in Conjunction with {MICCAI} 2019, Shenzhen,
  China, October 13 and 17, 2019, Proceedings}, volume 11851 of \emph{Lecture
  Notes in Computer Science}, pages 51--60. Springer, 2019.

\bibitem[Denker and LeCun(1991)]{denker1991transforming}
John~S. Denker and Yann LeCun.
\newblock {Transforming Neural-Net Output Levels to Probability Distributions}.
\newblock In R.~P. Lippmann, J.~E. Moody, and D.~S. Touretzky, editors,
  \emph{Advances in Neural Information Processing Systems 3}, pages 853--859.
  Morgan-Kaufmann, 1991.

\bibitem[Gal and Ghahramani(2016)]{gal2016dropout}
Yarin Gal and Zoubin Ghahramani.
\newblock {Dropout as a Bayesian Approximation: Representing Model Uncertainty
  in Deep Learning}.
\newblock In \emph{Proceedings of The 33rd International Conference on Machine
  Learning}, volume~48 of \emph{Proceedings of Machine Learning Research},
  pages 1050--1059, 2016.

\bibitem[Gustafsson et~al.(2020)Gustafsson, Danelljan, and
  Schon]{gustafsson2020evaluating}
F.~K. Gustafsson, M.~Danelljan, and T.~B. Schon.
\newblock {Evaluating Scalable Bayesian Deep Learning Methods for Robust
  Computer Vision}.
\newblock In \emph{2020 IEEE/CVF Conference on Computer Vision and Pattern
  Recognition Workshops (CVPRW)}, pages 1289--1298, Los Alamitos, CA, USA, jun
  2020. IEEE Computer Society.
\newblock \doi{10.1109/CVPRW50498.2020.00167}.
\newblock URL
  \url{https://doi.ieeecomputersociety.org/10.1109/CVPRW50498.2020.00167}.

\bibitem[Izmailov et~al.(2018)Izmailov, Podoprikhin, Garipov, Vetrov, and
  Wilson]{izmailov2018averaging}
Pavel Izmailov, Dmitrii Podoprikhin, Timur Garipov, Dmitry Vetrov, and
  Andrew~Gordon Wilson.
\newblock {Averaging Weights Leads to Wider Optima and Better Generalization}.
\newblock \emph{arXiv preprint arXiv:1803.05407}, 2018.

\bibitem[Jungo et~al.(2018)Jungo, Meier, Ermis, Blatti-Moreno, Herrmann, Wiest,
  and Reyes]{jungo2018effect}
Alain Jungo, Raphael Meier, Ekin Ermis, Marcela Blatti-Moreno, Evelyn Herrmann,
  Roland Wiest, and Mauricio Reyes.
\newblock {On the Effect of Inter-observer Variability for a Reliable
  Estimation of Uncertainty of Medical Image Segmentation}.
\newblock In Alejandro~F. Frangi, Julia~A. Schnabel, Christos Davatzikos,
  Carlos Alberola-L{\'o}pez, and Gabor Fichtinger, editors, \emph{Medical Image
  Computing and Computer Assisted Intervention -- MICCAI 2018}, pages 682--690,
  Cham, 2018. Springer International Publishing.
\newblock ISBN 978-3-030-00928-1.

\bibitem[Jungo et~al.(2020)Jungo, Balsiger, and Reyes]{jungo2020analyzing}
Alain Jungo, Fabian Balsiger, and Mauricio Reyes.
\newblock {Analyzing the Quality and Challenges of Uncertainty Estimations for
  Brain Tumor Segmentation}.
\newblock \emph{Frontiers in Neuroscience}, 14:\penalty0 282, 2020.
\newblock ISSN 1662-453X.
\newblock \doi{10.3389/fnins.2020.00282}.
\newblock URL
  \url{https://www.frontiersin.org/article/10.3389/fnins.2020.00282}.

\bibitem[Kendall and Gal(2017)]{kendall2017uncertainties}
Alex Kendall and Yarin Gal.
\newblock {What Uncertainties Do We Need in Bayesian Deep Learning for Computer
  Vision?}
\newblock In I.~Guyon, U.~V. Luxburg, S.~Bengio, H.~Wallach, R.~Fergus,
  S.~Vishwanathan, and R.~Garnett, editors, \emph{Advances in Neural
  Information Processing Systems}, volume~30, pages 5574--5584. Curran
  Associates, Inc., 2017.

\bibitem[Kendall et~al.(2017)Kendall, Badrinarayanan, and
  Cipolla]{kendall2015bayesian}
Alex Kendall, Vijay Badrinarayanan, and Roberto Cipolla.
\newblock {Bayesian SegNet: Model Uncertainty in Deep Convolutional
  Encoder-Decoder Architectures for Scene Understanding}.
\newblock In Gabriel~Brostow Tae-Kyun~Kim, Stefanos~Zafeiriou and Krystian
  Mikolajczyk, editors, \emph{Proceedings of the British Machine Vision
  Conference (BMVC)}, pages 57.1--57.12. BMVA Press, September 2017.
\newblock ISBN 1-901725-60-X.

\bibitem[Kiureghian and Ditlevsen(2009)]{kiureghian2009aleatory}
Armen~Der Kiureghian and Ove Ditlevsen.
\newblock Aleatory or epistemic? does it matter?
\newblock \emph{Structural Safety}, 31\penalty0 (2):\penalty0 105--112, 2009.
\newblock ISSN 0167-4730.
\newblock \doi{https://doi.org/10.1016/j.strusafe.2008.06.020}.
\newblock URL
  \url{https://www.sciencedirect.com/science/article/pii/S0167473008000556}.
\newblock Risk Acceptance and Risk Communication.

\bibitem[Lakshminarayanan et~al.(2016)Lakshminarayanan, Pritzel, and
  Blundell]{lakshminarayanan2016simple}
Balaji Lakshminarayanan, Alexander Pritzel, and Charles Blundell.
\newblock {Simple and scalable predictive uncertainty estimation using deep
  ensembles}.
\newblock \emph{arXiv preprint arXiv:1612.01474}, 2016.

\bibitem[Laves et~al.(2020)Laves, Ihler, Fast, Kahrs, and
  Ortmaier]{laves2020well}
Max-Heinrich Laves, Sontje Ihler, Jacob~F Fast, L{\"u}der~A Kahrs, and Tobias
  Ortmaier.
\newblock {Well-Calibrated Regression Uncertainty in Medical Imaging with Deep
  Learning}.
\newblock In \emph{Medical Imaging with Deep Learning}, 2020.

\bibitem[Luo et~al.(2019)Luo, Liu, Tong, Jiang, Yuan, Zhao, and
  Shang]{luo2019carotid}
Lian Luo, Shuai Liu, Xinyu Tong, Peirong Jiang, Chun Yuan, Xihai Zhao, and Fei
  Shang.
\newblock {Carotid artery segmentation using level set method with double
  adaptive threshold (DATLS) on TOF-MRA images}.
\newblock \emph{Magnetic Resonance Imaging}, 63:\penalty0 123--130, 2019.
\newblock ISSN 0730-725X.
\newblock \doi{https://doi.org/10.1016/j.mri.2019.08.002}.
\newblock URL
  \url{https://www.sciencedirect.com/science/article/pii/S0730725X19301791}.

\bibitem[MacKay(1992)]{mackay1992practical}
David J.~C. MacKay.
\newblock {A Practical Bayesian Framework for Backpropagation Networks}.
\newblock \emph{Neural Computation}, 4\penalty0 (3):\penalty0 448--472, 1992.
\newblock URL \url{https://doi.org/10.1162/neco.1992.4.3.448}.

\bibitem[Maddox et~al.(2019)Maddox, Izmailov, Garipov, Vetrov, and
  Wilson]{maddox2019simple}
Wesley~J Maddox, Pavel Izmailov, Timur Garipov, Dmitry~P Vetrov, and
  Andrew~Gordon Wilson.
\newblock {A Simple Baseline for Bayesian Uncertainty in Deep Learning}.
\newblock In H.~Wallach, H.~Larochelle, A.~Beygelzimer, F.~d\textquotesingle
  Alch\'{e}-Buc, E.~Fox, and R.~Garnett, editors, \emph{Advances in Neural
  Information Processing Systems}, volume~32. Curran Associates, Inc., 2019.
\newblock URL
  \url{https://proceedings.neurips.cc/paper/2019/file/118921efba23fc329e6560b27861f0c2-Paper.pdf}.

\bibitem[Makowski et~al.(2019)Makowski, Ben-Shachar, and
  L{\"u}decke]{makowski2019bayestestr}
Dominique Makowski, Mattan Ben-Shachar, and Daniel L{\"u}decke.
\newblock {bayestestR: Describing Effects and their Uncertainty, Existence and
  Significance within the Bayesian Framework}.
\newblock \emph{Journal of Open Source Software}, 4\penalty0 (40):\penalty0
  1541, 2019.

\bibitem[McElreath(2020)]{mcelreath2020statistical}
Richard McElreath.
\newblock \emph{{Statistical Rethinking: A Bayesian Course with Examples in R
  and Stan}}.
\newblock CRC press, 2020.

\bibitem[Mehrtash et~al.(2019)Mehrtash, Wells~III, Tempany, Abolmaesumi, and
  Kapur]{mehrtash2019confidence}
Alireza Mehrtash, William~M Wells~III, Clare~M Tempany, Purang Abolmaesumi, and
  Tina Kapur.
\newblock {Confidence Calibration and Predictive Uncertainty Estimation for
  Deep Medical Image Segmentation}.
\newblock \emph{arXiv preprint arXiv:1911.13273}, 2019.

\bibitem[Mehta et~al.(2020)Mehta, Filos, Gal, and Arbel]{mehta2020uncertainty}
Raghav Mehta, Angelos Filos, Yarin Gal, and Tal Arbel.
\newblock {Uncertainty Evaluation Metrics for Brain Tumour Segmentation}.
\newblock In \emph{Medical Imaging with Deep Learning}, 2020.

\bibitem[Menze et~al.(2014)Menze, Jakab, Bauer, Kalpathy-Cramer, Farahani,
  Kirby, Burren, Porz, Slotboom, Wiest, et~al.]{menze2014multimodal}
Bjoern~H Menze, Andras Jakab, Stefan Bauer, Jayashree Kalpathy-Cramer, Keyvan
  Farahani, Justin Kirby, Yuliya Burren, Nicole Porz, Johannes Slotboom, Roland
  Wiest, et~al.
\newblock {The Multimodal Brain Tumor Image Segmentation Benchmark (BRATS)}.
\newblock \emph{IEEE transactions on medical imaging}, 34\penalty0
  (10):\penalty0 1993--2024, 2014.

\bibitem[Michelmore et~al.(2018)Michelmore, Kwiatkowska, and
  Gal]{michelmore2018evaluating}
Rhiannon Michelmore, Marta Kwiatkowska, and Yarin Gal.
\newblock {Evaluating Uncertainty Quantification in End-to-End Autonomous
  Driving Control}.
\newblock \emph{arXiv preprint arXiv:1811.06817}, 2018.

\bibitem[Milletari et~al.(2016)Milletari, Navab, and Ahmadi]{milletari2016v}
Fausto Milletari, Nassir Navab, and Seyed-Ahmad Ahmadi.
\newblock {V-net: Fully Convolutional Neural Networks For Volumetric Medical
  Image Segmentation}.
\newblock In \emph{2016 Fourth International Conference on 3D Vision (3DV)},
  pages 565--571. IEEE, 2016.

\bibitem[Mobiny et~al.(2019{\natexlab{a}})Mobiny, Nguyen, Moulik, Garg, and
  Wu]{mobiny2019dropconnect}
Aryan Mobiny, Hien~V Nguyen, Supratik Moulik, Naveen Garg, and Carol~C Wu.
\newblock {DropConnect Is Effective in Modeling Uncertainty of Bayesian Deep
  Networks}.
\newblock \emph{arXiv preprint arXiv:1906.04569}, 2019{\natexlab{a}}.

\bibitem[Mobiny et~al.(2019{\natexlab{b}})Mobiny, Singh, and
  Van~Nguyen]{mobiny2019risk}
Aryan Mobiny, Aditi Singh, and Hien Van~Nguyen.
\newblock {Risk-Aware Machine Learning Classifier for Skin Lesion Diagnosis}.
\newblock \emph{Journal of clinical medicine}, 8\penalty0 (8):\penalty0 1241,
  2019{\natexlab{b}}.

\bibitem[Mukhoti and Gal(2018)]{mukhoti2018evaluating}
Jishnu Mukhoti and Yarin Gal.
\newblock {Evaluating bayesian deep learning methods for semantic
  segmentation}.
\newblock \emph{arXiv preprint arXiv:1811.12709}, 2018.

\bibitem[Nair et~al.(2020)Nair, Precup, Arnold, and Arbel]{nair2020exploring}
Tanya Nair, Doina Precup, Douglas~L. Arnold, and Tal Arbel.
\newblock {Exploring uncertainty measures in deep networks for Multiple
  sclerosis lesion detection and segmentation}.
\newblock \emph{Medical Image Analysis}, 59:\penalty0 101557, 2020.
\newblock ISSN 1361-8415.

\bibitem[Neal(1993)]{neal1993bayesian}
Radford Neal.
\newblock {Bayesian Learning via Stochastic Dynamics}.
\newblock In S.~Hanson, J.~Cowan, and C.~Giles, editors, \emph{Advances in
  Neural Information Processing Systems}, volume~5, pages 475--482.
  Morgan-Kaufmann, 1993.

\bibitem[Orlando et~al.(2019)Orlando, Seeb{\"o}ck, Bogunovi{\'c}, Klimscha,
  Grechenig, Waldstein, Gerendas, and Schmidt-Erfurth]{orlando2019u2}
Jos{\'e}~Ignacio Orlando, Philipp Seeb{\"o}ck, Hrvoje Bogunovi{\'c}, Sophie
  Klimscha, Christoph Grechenig, Sebastian Waldstein, Bianca~S Gerendas, and
  Ursula Schmidt-Erfurth.
\newblock {U2-Net: A Bayesian U-Net Model with Epistemic Uncertainty Feedback
  for Photoreceptor Layer Segmentation in Pathological OCT Scans}.
\newblock In \emph{2019 IEEE 16th International Symposium on Biomedical Imaging
  (ISBI 2019)}, pages 1441--1445. IEEE, 2019.

\bibitem[Paszke et~al.(2019)Paszke, Gross, Massa, Lerer, Bradbury, Chanan,
  Killeen, Lin, Gimelshein, Antiga, et~al.]{paszke2019pytorch}
Adam Paszke, Sam Gross, Francisco Massa, Adam Lerer, James Bradbury, Gregory
  Chanan, Trevor Killeen, Zeming Lin, Natalia Gimelshein, Luca Antiga, et~al.
\newblock {PyTorch: An imperative style, high-performance deep learning
  library}.
\newblock In \emph{Advances in Neural Information Processing Systems}, pages
  8024--8035, 2019.

\bibitem[Pedregosa et~al.(2011)Pedregosa, Varoquaux, Gramfort, Michel, Thirion,
  Grisel, Blondel, Prettenhofer, Weiss, Dubourg, Vanderplas, Passos,
  Cournapeau, Brucher, Perrot, and Duchesnay]{pedregosa2011scikit}
Fabian Pedregosa, Gaël Varoquaux, Alexandre Gramfort, Vincent Michel, Bertrand
  Thirion, Olivier Grisel, Mathieu Blondel, Peter Prettenhofer, Ron Weiss,
  Vincent Dubourg, Jake Vanderplas, Alexandre Passos, David Cournapeau,
  Matthieu Brucher, Matthieu Perrot, and Edouard Duchesnay.
\newblock {Scikit-learn: Machine Learning in Python}.
\newblock \emph{Journal of Machine Learning Research}, 12:\penalty0 2825--2830,
  2011.

\bibitem[Ronneberger et~al.(2015)Ronneberger, Fischer, and
  Brox]{ronneberger2015u}
Olaf Ronneberger, Philipp Fischer, and Thomas Brox.
\newblock {U-Net: Convolutional Networks for Biomedical Image Segmentation}.
\newblock In Nassir Navab, Joachim Hornegger, William~M. Wells, and
  Alejandro~F. Frangi, editors, \emph{Medical Image Computing and
  Computer-Assisted Intervention -- MICCAI 2015}, pages 234--241, Cham, 2015.
  Springer International Publishing.
\newblock ISBN 978-3-319-24574-4.

\bibitem[Sedghi et~al.(2019)Sedghi, Kapur, Luo, Mousavi, and
  Wells]{sedghi2019probabilistic}
Alireza Sedghi, Tina Kapur, Jie Luo, Parvin Mousavi, and William~M Wells.
\newblock {Probabilistic Image Registration via Deep Multi-class
  Classification: Characterizing Uncertainty}.
\newblock In \emph{Uncertainty for Safe Utilization of Machine Learning in
  Medical Imaging and Clinical Image-Based Procedures}, pages 12--22. Springer,
  2019.

\bibitem[Seeb{\"o}ck et~al.(2019)Seeb{\"o}ck, Orlando, Schlegl, Waldstein,
  Bogunovi{\'c}, Klimscha, Langs, and Schmidt-Erfurth]{seebock2019exploiting}
Philipp Seeb{\"o}ck, Jos{\'e}~Ignacio Orlando, Thomas Schlegl, Sebastian~M
  Waldstein, Hrvoje Bogunovi{\'c}, Sophie Klimscha, Georg Langs, and Ursula
  Schmidt-Erfurth.
\newblock {Exploiting Epistemic Uncertainty of Anatomy Segmentation for Anomaly
  Detection in Retinal OCT}.
\newblock \emph{IEEE transactions on medical imaging}, 39\penalty0
  (1):\penalty0 87--98, 2019.

\bibitem[Srivastava et~al.(2014)Srivastava, Hinton, Krizhevsky, Sutskever, and
  Salakhutdinov]{srivastava2014dropout}
Nitish Srivastava, Geoffrey Hinton, Alex Krizhevsky, Ilya Sutskever, and Ruslan
  Salakhutdinov.
\newblock {Dropout: A Simple Way to Prevent Neural Networks from Overfitting}.
\newblock \emph{J. Mach. Learn. Res.}, 15\penalty0 (1):\penalty0 1929–1958,
  January 2014.
\newblock ISSN 1532-4435.

\bibitem[Taghanaki et~al.(2021)Taghanaki, Abhishek, Cohen, Cohen-Adad, and
  Hamarneh]{taghanaki2021deep}
Saeid~Asgari Taghanaki, Kumar Abhishek, Joseph~Paul Cohen, Julien Cohen-Adad,
  and Ghassan Hamarneh.
\newblock {Deep semantic segmentation of natural and medical images: A review}.
\newblock \emph{Artificial Intelligence Review}, 54\penalty0 (1):\penalty0
  137--178, 2021.

\bibitem[Teye et~al.(2018)Teye, Azizpour, and Smith]{teye2018bayesian}
Mattias Teye, Hossein Azizpour, and Kevin Smith.
\newblock Bayesian uncertainty estimation for batch normalized deep networks.
\newblock In \emph{International Conference on Machine Learning}, pages
  4907--4916. PMLR, 2018.

\bibitem[Truijman et~al.(2014)Truijman, Kooi, Van~Dijk, de~Rotte, van~der Kolk,
  Liem, Schreuder, Boersma, Mess, van Oostenbrugge, et~al.]{truijman2014plaque}
MTB Truijman, M~Eline Kooi, AC~Van~Dijk, AAJ de~Rotte, AG~van~der Kolk,
  MI~Liem, FHBM Schreuder, Elly Boersma, WH~Mess, Robert~Jan van Oostenbrugge,
  et~al.
\newblock {Plaque At RISK (PARISK): prospective multicenter study to improve
  diagnosis of high-risk carotid plaques}.
\newblock \emph{International Journal of Stroke}, 9\penalty0 (6):\penalty0
  747--754, 2014.

\bibitem[Van~Molle et~al.(2019)Van~Molle, Verbelen, De~Boom, Vankeirsbilck,
  De~Vylder, Diricx, Kimpe, Simoens, and Dhoedt]{de2019quantifying}
Pieter Van~Molle, Tim Verbelen, Cedric De~Boom, Bert Vankeirsbilck, Jonas
  De~Vylder, Bart Diricx, Tom Kimpe, Pieter Simoens, and Bart Dhoedt.
\newblock {Quantifying Uncertainty of Deep Neural Networks in Skin Lesion
  Classification}.
\newblock In Hayit Greenspan, Ryutaro Tanno, Marius Erdt, Tal Arbel, Christian
  Baumgartner, Adrian Dalca, Carole~H. Sudre, William~M. Wells, Klaus
  Drechsler, Marius~George Linguraru, Cristina Oyarzun~Laura, Raj Shekhar,
  Stefan Wesarg, and Miguel~{\'A}ngel Gonz{\'a}lez~Ballester, editors,
  \emph{Uncertainty for Safe Utilization of Machine Learning in Medical Imaging
  and Clinical Image-Based Procedures}, pages 52--61, Cham, 2019. Springer
  International Publishing.
\newblock ISBN 978-3-030-32689-0.

\bibitem[Wang et~al.(2019)Wang, Li, Aertsen, Deprest, Ourselin, and
  Vercauteren]{wang2019aleatoric}
Guotai Wang, Wenqi Li, Michael Aertsen, Jan Deprest, Sébastien Ourselin, and
  Tom Vercauteren.
\newblock {Aleatoric uncertainty estimation with test-time augmentation for
  medical image segmentation with convolutional neural networks}.
\newblock \emph{Neurocomputing}, 338:\penalty0 34 -- 45, 2019.
\newblock ISSN 0925-2312.

\bibitem[Welling and Teh(2011)]{welling2011bayesian}
Max Welling and Yee~W Teh.
\newblock {Bayesian learning via stochastic gradient Langevin dynamics}.
\newblock In \emph{Proceedings of the 28th international conference on machine
  learning (ICML-11)}, pages 681--688. Citeseer, 2011.

\bibitem[Wilson and Izmailov(2020)]{wilson2020bayesian}
Andrew~Gordon Wilson and Pavel Izmailov.
\newblock {Bayesian deep learning and a probabilistic perspective of
  generalization}.
\newblock \emph{arXiv preprint arXiv:2002.08791}, 2020.

\bibitem[{World Health Organization}(2011)]{world2011global}
{World Health Organization}.
\newblock {Global atlas on cardiovascular disease prevention and control:
  published by the World Health Organization in collaboration with the World
  Heart Federation and the World Stroke Organization}.
\newblock 2011.

\bibitem[{World Health Organization}(2014)]{world2014global}
{World Health Organization}.
\newblock \emph{{Global status report on noncommunicable diseases 2014}}.
\newblock Number WHO/NMH/NVI/15.1. World Health Organization, 2014.

\bibitem[Wu et~al.(2019)Wu, Xin, Yang, Sun, Xu, Zheng, and Yuan]{wu2019deep}
Jiayi Wu, Jingmin Xin, Xiaofeng Yang, Jie Sun, Dongxiang Xu, Nanning Zheng, and
  Chun Yuan.
\newblock {Deep morphology aided diagnosis network for segmentation of carotid
  artery vessel wall and diagnosis of carotid atherosclerosis on black-blood
  vessel wall MRI}.
\newblock \emph{Medical physics}, 46\penalty0 (12):\penalty0 5544--5561, 2019.

\bibitem[Zeiler(2012)]{zeiler2012adadelta}
Matthew~D Zeiler.
\newblock {ADADELTA: An Adaptive Learning Rate Method}.
\newblock \emph{arXiv preprint arXiv:1212.5701}, 2012.

\bibitem[Zhu et~al.(2021)Zhu, Wang, Teng, Chen, Huang, Xia, Mao, and
  Bai]{zhu2021cascaded}
Chenglu Zhu, Xiaoyan Wang, Zhongzhao Teng, Shengyong Chen, Xiaojie Huang, Ming
  Xia, Lizhao Mao, and Cong Bai.
\newblock {Cascaded residual U-net for fully automatic segmentation of 3D
  carotid artery in high-resolution multi-contrast MR images}.
\newblock \emph{Physics in Medicine \& Biology}, 66\penalty0 (4):\penalty0
  045033, 2021.

\end{thebibliography}

\newpage

\appendix

\section*{Appendix A: Dataset description}

\begin{table}[H]
\centering
\caption{MR images scan parameters. (QIR = quadruple inversion recovery, TSE = turbo spin echo, IR = inversion recovery, FFE = fast field echo and, TFE = turbo field echo, FA = flip angle, AVS = acquired voxel size, RVS = reconstructed voxel size)}
\label{tab:mriconfig}
\begin{tabular}{l|l|l|l|l|l}
\hline
Pulse                         & \multicolumn{2}{l|}{T1wQIR TSE} & TOF FFE     & IR-TFE      & T2w TSE     \\ \cline{2-3}
Sequence                      & pre-contrast   & post-contrast  &             &             &             \\ \hline
Repetion time ($ms$)            & 800            & 800            & 20          & 3.3         & 4800        \\
Echo time ($ms$)                & 10             & 10             & 5           & 2.1         & 49          \\
Inversion time ($ms$)                & 282,61             & 282,61            &            & 304         &           \\
FA (degrees)                        & 90              & 90             & 20          & 15          & 90           \\
AVS ($mm^2$) & 0.62 x 0.67    & 0.62 x 0.67    & 0.62 x 0.62 & 0.62 x 0.63 & 0.62 x 0.63 \\
RVS ($mm^2$) & 0.30 x 0.30    & 0.30 x 0.30    & 0.30 x 0.30 & 0.30 x 0.24 & 0.30 x 0.30 \\
Slice thickness ($mm$)          & 2              & 2              & 2           & 2           & 2           \\ \hline
\end{tabular}
\end{table}

\begin{figure}[H]
    \centering
    \begin{subfigure}[b]{0.48\textwidth}
        \includegraphics[width=\textwidth]{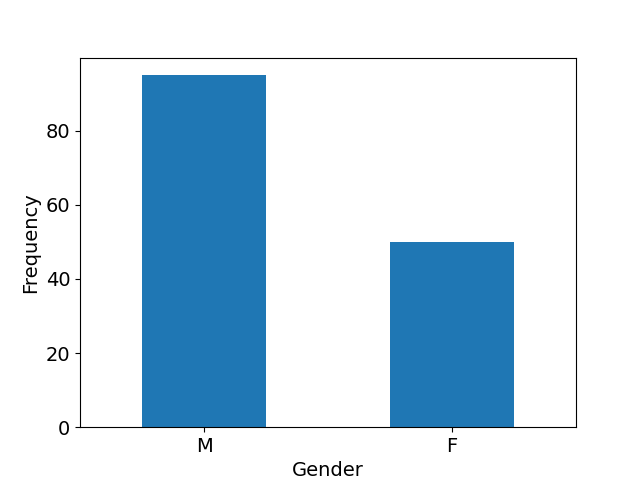}
    \end{subfigure}
    \begin{subfigure}[b]{0.48\textwidth}
        \includegraphics[width=\textwidth]{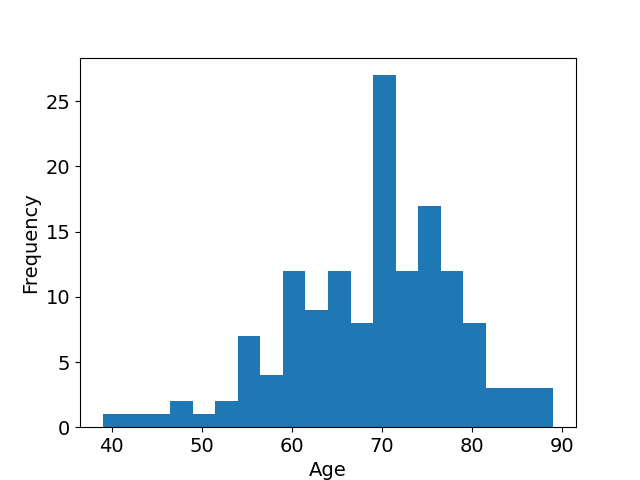}
    \end{subfigure}
    \caption{Description of the dataset. The left figure corresponds to the distribution of gender across the 145 patients under study and the right figure corresponds to the distribution of age in years across the 145 patients under study ($\mu = 68.8$, $\sigma = 9.2$).}
    \label{fig:dataset_statistics}
\end{figure}

\section*{Appendix B: Network structure}

\begin{figure}[H]
    \centering
    \def\svgwidth{0.75\columnwidth}
    \tiny 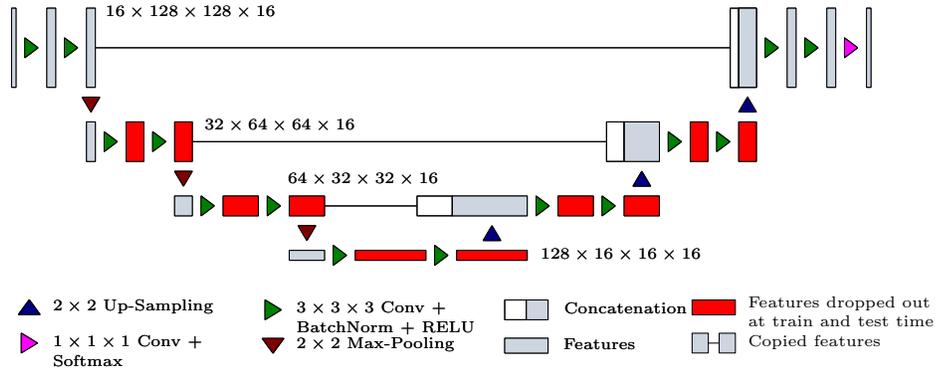
    \caption{Network structure}
    \label{fig:network}
\end{figure}

\section*{Appendix C: Posterior distribution of $p_{A > B}$ for the combined AUC-PR performance measure}

\begin{figure}[H]
    \centering
    \foreach \metrica in {Averaged-Entropy, Averaged-Variance,  Multi-class-Entropy, Multi-class-MI, Similarity-BC}{
        \foreach \metricb in {Averaged-Variance,  Multi-class-Entropy, Multi-class-MI, Similarity-BC, Similarity-KL}{
  		\begin{subfigure}[b]{0.18\textwidth}
		    \includegraphics[width=\textwidth]{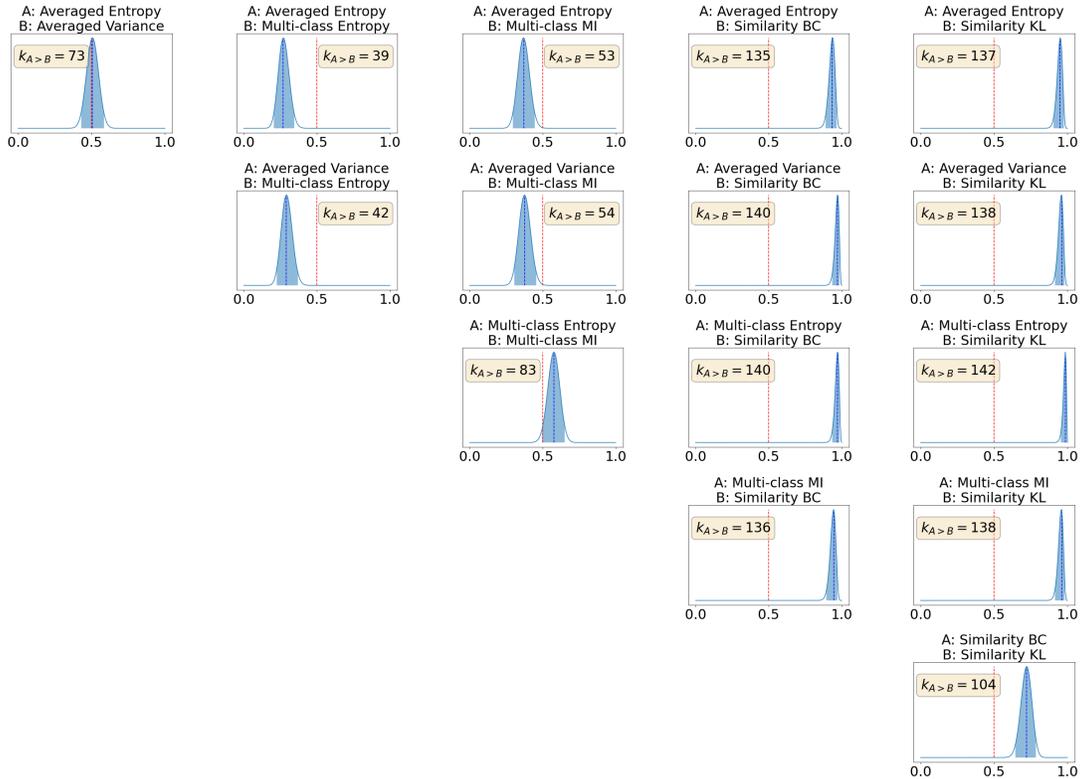}
		\end{subfigure}
        } \\
    }
    \caption{Posterior distribution of $p_{A > B}$ for the combined AUC-PR performance measure and the combined uncertainty maps pairs A and B, the red dashed line represents the expected value if compared uncertainty maps perform equally and the blue area under the curve represents the 95\% credible interval}
    \label{fig:statsignmulticlass}
\end{figure}

\section*{Appendix D: Posterior distribution of $p_{A > B}$ for the class-specific AUC-PR performance measure}

\begin{figure}[H]
    \centering
    \foreach \metrica in {Class-wise-Entropy, Class-wise-Variance, One-vs-all-Entropy}{
        \foreach \metricb in {Class-wise-Variance, One-vs-all-Entropy, One-vs-all-MI}{
		\begin{subfigure}[b]{0.20\textwidth}
		    \includegraphics[width=\textwidth]{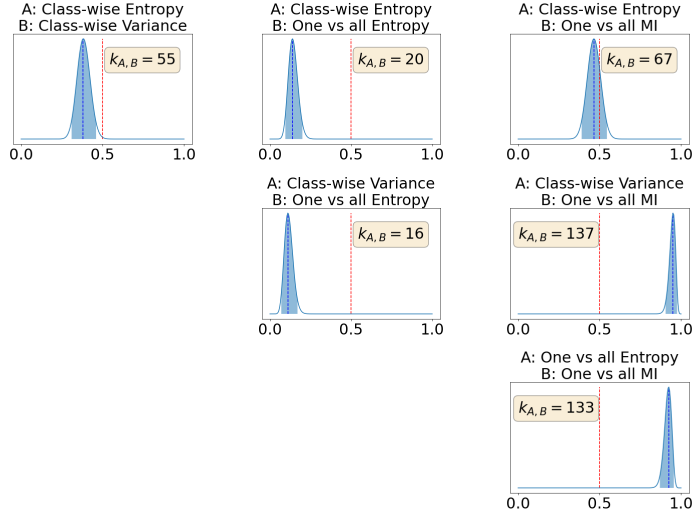}
		\end{subfigure}
        }\\
    }
    \caption{Posterior distribution of $p_{A > B}$ for the class-specific AUC-PR performance measure and the background uncertainty maps pairs A and B, the red dashed line represents the expected value if compared uncertainty maps perform equally and the blue area under the curve represents the 95\% credible interval}
    \label{fig:statsignaucprbackground}
\end{figure}

\begin{figure}[H]
    \centering
    \foreach \metrica in {Class-wise-Entropy, Class-wise-Variance, One-vs-all-Entropy}{
        \foreach \metricb in {Class-wise-Variance, One-vs-all-Entropy, One-vs-all-MI}{
		\begin{subfigure}[b]{0.20\textwidth}
		    \includegraphics[width=\textwidth]{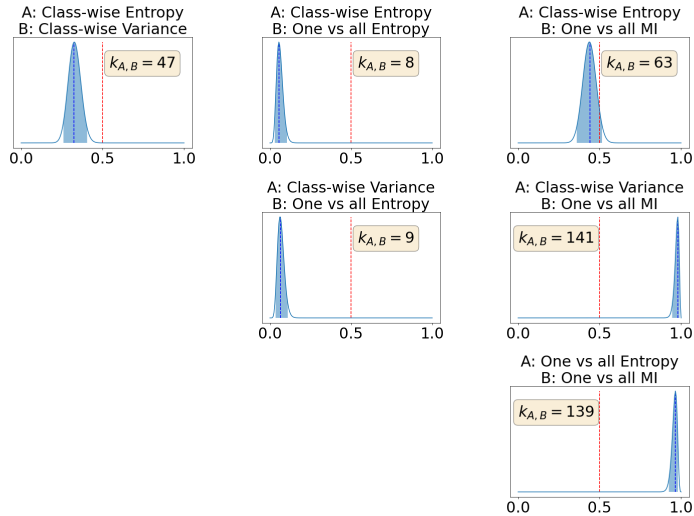}
		\end{subfigure}
        }\\
    }
    \caption{Posterior distribution of $p_{A > B}$ for the class-specific AUC-PR performance measure and the vessel wall uncertainty maps pairs A and B, the red dashed line represents the expected value if compared uncertainty maps perform equally and the blue area under the curve represents the 95\% credible interval}
    \label{fig:statsignaucprvesselwall}
\end{figure}

\begin{figure}[H]
    \centering
    \foreach \metrica in {Class-wise-Entropy, Class-wise-Variance, One-vs-all-Entropy}{
        \foreach \metricb in {Class-wise-Variance, One-vs-all-Entropy, One-vs-all-MI}{
		\begin{subfigure}[b]{0.20\textwidth}
		    \includegraphics[width=\textwidth]{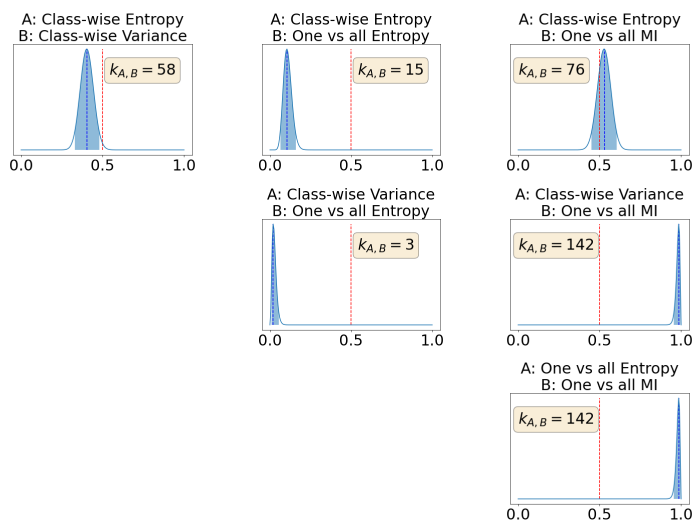}
		\end{subfigure}
        }\\
    }
    \caption{Posterior distribution of $p_{A > B}$ for AUC-PR performance measure and the lumen uncertainty maps pairs A and B, the red dashed line represents the expected value if compared uncertainty maps perform equally and the blue area under the curve represents the 95\% credible interval}
    \label{fig:statsignaucprlumen}
\end{figure}

\section*{Appendix E: Posterior distribution of $p_{A > B}$ for the BRATS-UNC performance measure}

\begin{figure}[H]
    \centering
    \foreach \metrica in {Class-wise-Entropy, Class-wise-Variance, One-vs-all-Entropy}{
        \foreach \metricb in {Class-wise-Variance, One-vs-all-Entropy, One-vs-all-MI}{
		\begin{subfigure}[b]{0.20\textwidth}
		    \includegraphics[width=\textwidth]{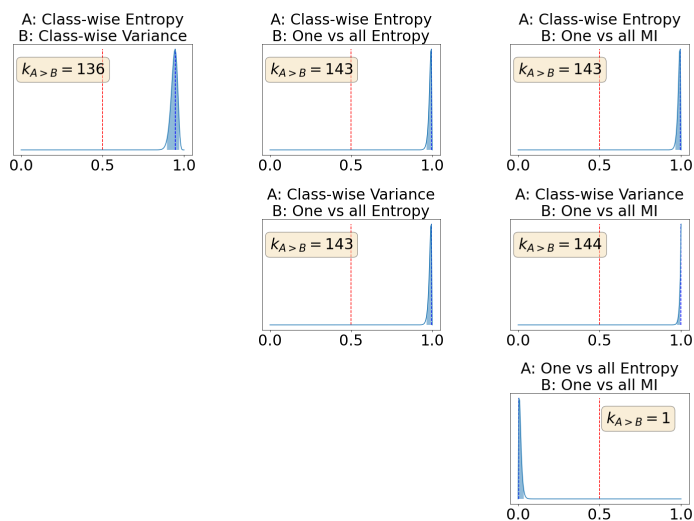}
		\end{subfigure}
        }\\
    }
    \caption{Posterior distribution of $p_{A > B}$ for the BRATS-UNC performance measure and the background uncertainty maps pairs A and B, the red dashed line represents the expected value if compared uncertainty maps perform equally and the blue area under the curve represents the 95\% credible interval}
    \label{fig:statsignbackground}
\end{figure}

\begin{figure}[H]
    \centering
    \foreach \metrica in {Class-wise-Entropy, Class-wise-Variance, One-vs-all-Entropy}{
        \foreach \metricb in {Class-wise-Variance, One-vs-all-Entropy, One-vs-all-MI}{
		\begin{subfigure}[b]{0.20\textwidth}
		    \includegraphics[width=\textwidth]{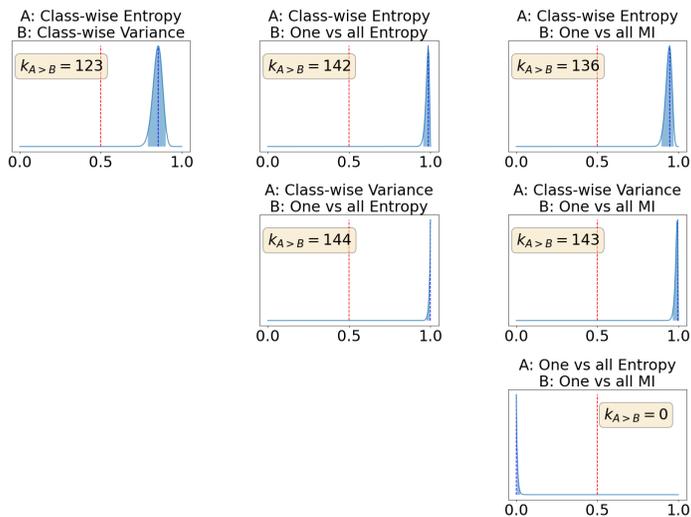}
		\end{subfigure}
        }\\
    }
    \caption{Posterior distribution of $p_{A > B}$ for the BRATS-UNC performance measures and the vessel wall uncertainty maps pairs A and B, the red dashed line represents the expected value if compared uncertainty maps perform equally and the blue area under the curve represents the 95\% credible interval}
    \label{fig:statsignvesselwall}
\end{figure}

\begin{figure}[H]
    \centering
    \foreach \metrica in {Class-wise-Entropy, Class-wise-Variance, One-vs-all-Entropy}{
        \foreach \metricb in {Class-wise-Variance, One-vs-all-Entropy, One-vs-all-MI}{
		\begin{subfigure}[b]{0.20\textwidth}
		    \includegraphics[width=\textwidth]{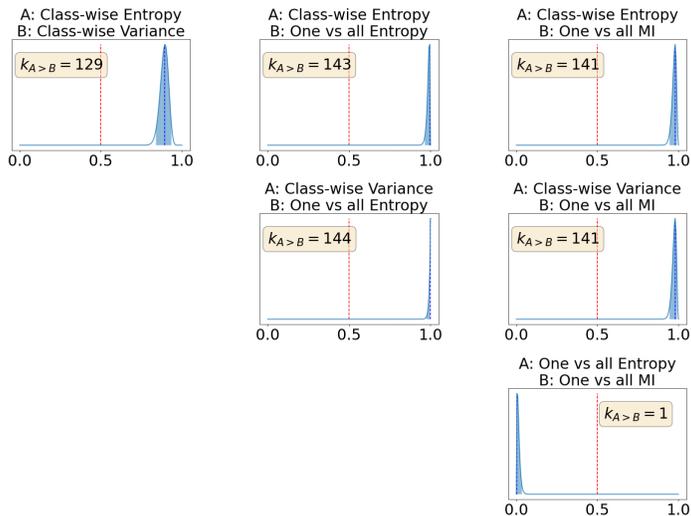}
		\end{subfigure}
        }\\
    }
    \caption{Posterior distribution of $p_{A > B}$ for the BRATS-UNC performance measures and the lumen uncertainty maps pairs A and B, the red dashed line represents the expected value if compared uncertainty maps perform equally and the blue area under the curve represents the 95\% credible interval}
    \label{fig:statsignlumen}
\end{figure}

\section*{Appendix F: $p_{A > B}$ correlations}

\begin{figure}[H]
    \centering
    \foreach \metric in {AUCPRMetric, AUCPRClassWiseMetric_0, AUCPRClassWiseMetric_1, AUCPRClassWiseMetric_2, BRATSMetric_0, BRATSMetric_1, BRATSMetric_2}
    {
        \foreach \subset in {AMC, MUMC, UMCU, classic, gaussian}{
                \begin{subfigure}[b]{0.17\textwidth}
                    \includegraphics[width=\textwidth]{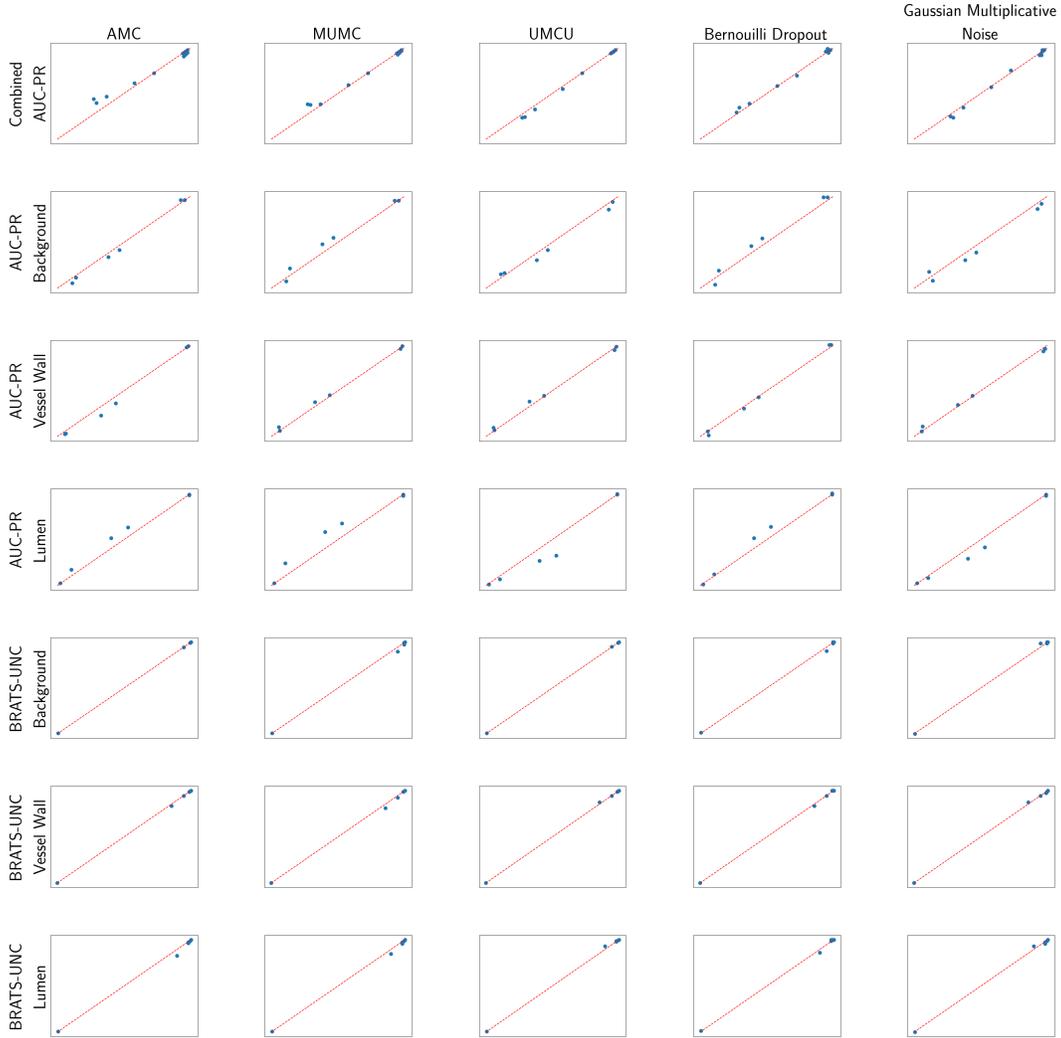}
                \end{subfigure}
        }
    }\\
    \caption{
        Correlation plots between the mode of the posterior distribution of $p_{A > B}$ computed on the whole results and the mode of the posterior distribution of $p_{A > B}$ computed on a subset of the results. Each plot contains blue dots that represent all the pairwise comparisons between uncertainty maps A and B for a given subset of the results (indicated in the column header) and a given performance measure (indicated in the row header). The x-axis correspond to the mode of the posterior distribution of $p_{A > B}$ observed on the subset of the results and the y-axis correspond to the mode of the posterior distribution of $p_{A > B}$ observed on the whole results. The rows correspond to the different performance measure (Combined AUC-PR, AUC-PR Background, AUC-PR Vessel Wall , AUC-PR Lumen, BRATS-UNC Background, BRATS-UNC Vessel Wall and BRATS-UNC Lumen). The three first columns correspond to the different the different test sets (AMC, MUMC and UMCU). The two last columns correspond to the experiment for the different type of dropout (Bernoulli dropout or Gaussian multiplicative noise).
    }
    \label{fig:center_correlation}
\end{figure}

\section*{Appendix G: Best model averaged curves}

\begin{figure}[H]
    \centering
    \includegraphics[width=.6\textwidth]{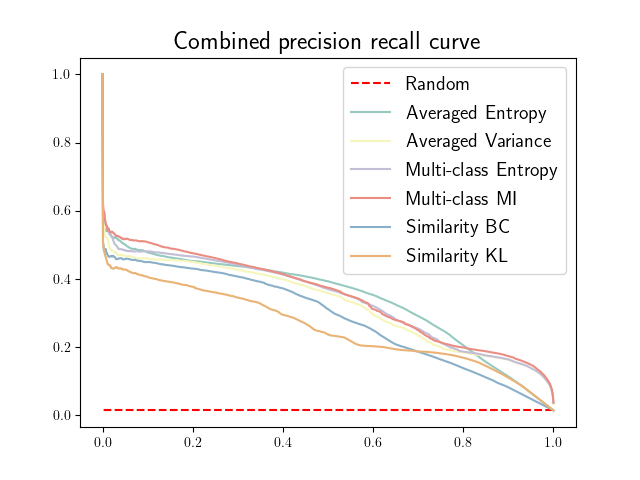}
    \caption{Combined precision recall curve of the best model averaged over the test set. The dashed line correspond to the random behaviour.}
    \label{fig:combined_pr_curve}
\end{figure}

\begin{figure}[H]
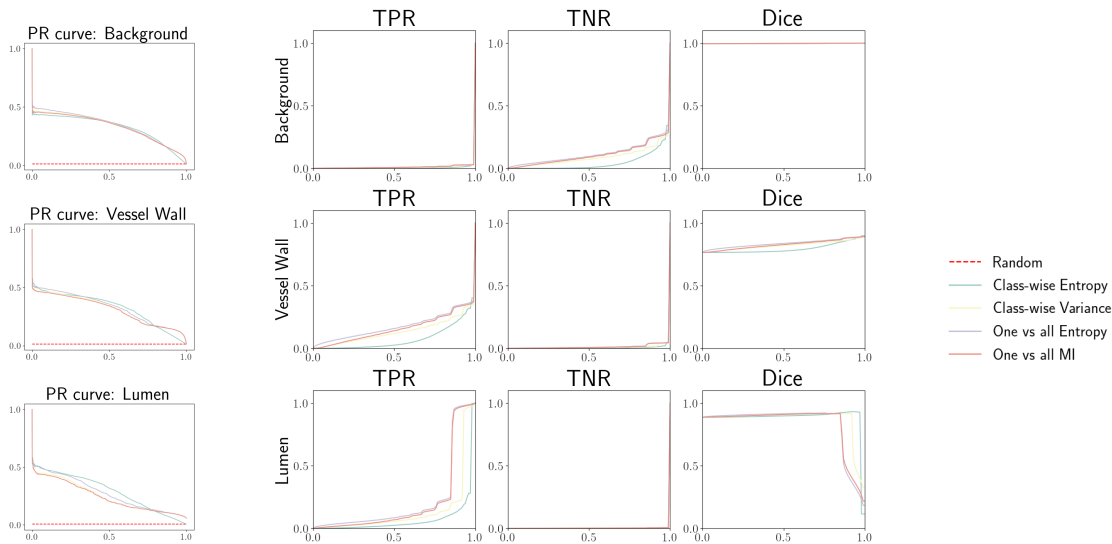

    \centering
    \foreach \class in {Background, Vessel_Wall, Lumen}
    {

        \begin{subfigure}[b]{.19\textwidth}
            \includegraphics[width=\textwidth]{figures/plot_pr_curves/158_pr_curve_\class.png}
        \end{subfigure}
        \begin{subfigure}[b]{.62\textwidth}
            \includegraphics[width=\textwidth]{figures/plot_brats_curves/158_brats_curves_\class.png}
        \end{subfigure}
        \begin{subfigure}[b]{.15\textwidth}
            \includegraphics[width=\textwidth]{figures/plot_pr_curves/legend_\class.png}
        \end{subfigure}
    }
    \caption{The three plots on the left correspond to the class-specific precision recall curves of the best model averaged over the test set. On the right the curves correspond to the 3 components integrated in the BRATS-UNC performance measure of the best model averaged over the test set. All plots have a shared legend on the right.}
    \label{fig:curves}
\end{figure}

\section*{Appendix H: Patient-wise $p_{A > B}$ of the 10 best models}

\begin{table}[H]
    \caption{95\% credible interval of the posterior distribution of patient-wise $p_{A > B}$ of the 10 best models for the combined AUC-PR performance measure and the combined uncertainty maps pairs A and B. The rows correspond to the uncertainty maps A and the columns to the uncertainty map B. The statistically significant differences are reported in bold ($0.5 \notin I_{95\%}$). $p_{A > B} > 0.5$ means that uncertainty map A is better than uncertainty map B and the contrary if $p_{A > B} < 0.5$ (Av = Averaged, Mu = Multi-class, Sim = Similarity)}
    \label{tab:top10multiclasssignificance}
    \centering
    \begin{tabular}{p{0.14\textwidth}|p{0.14\textwidth}|p{0.14\textwidth}|p{0.14\textwidth}|p{0.14\textwidth}|p{0.14\textwidth}p{0.14\textwidth}}
\hline
{} &       AV Variance &     Mu Entropy &         Mu MI &          Sim BC &           Sim KL \\
\hline
AV Entropy    &  $\textbf{[0.23, 0.48]}$ &  $\textbf{[0.07, 0.25]}$ &  $\textbf{[0.09, 0.30]}$ &   $\textbf{[0.90, 1.00]}$ &    $\textbf{[0.90, 1.00]}$ \\
AV Variance   &                         &    $\textbf{[0.00, 0.10]}$ &         $[0.26, 0.52]$ &  $\textbf{[0.93, 1.00]}$ &   $\textbf{[0.93, 1.00]}$ \\
Mu Entropy &                         &                         &         $[0.42, 0.68]$ &  $\textbf{[0.93, 1.00]}$ &   $\textbf{[0.93, 1.00]}$ \\
Mu MI      &                         &                         &                        &  $\textbf{[0.93, 1.00]}$ &   $\textbf{[0.93, 1.00]}$ \\
Sim BC       &                         &                         &                        &                        &  $\textbf{[0.58, 0.82]}$ \\
\hline
\end{tabular}

\end{table}

\begin{table}[H]
\caption{95\% credible interval of the posterior distribution of patient-wise $p_{A > B}$ of the 10 best models for the class-specific AUC-PR performance measure and the class-specific uncertainty maps pairs A and B. The rows correspond to the uncertainty maps A and the columns to the uncertainty map B. The statistically significant results are reported in bold ($0.5 \notin I_{95\%}$). $p_{A, B} > 0.5$ means that uncertainty map A is better than uncertainty map B and the contrary if $p_{A > B} < 0.5$. (CW = Class-wise, 1vA = One versus all)}
    \label{tab:top10classwiseaucprsignificance}
    \centering
    \subcaption*{Background}
    \begin{tabular}{p{0.167\textwidth}|p{0.167\textwidth}|p{0.167\textwidth}|p{0.167\textwidth}p{0.167\textwidth}}
\hline
{} &    CW Variance &      1vA Entropy &           1vA MI \\
\hline
CW Entropy  &  $\textbf{[0.17, 0.40]}$ &  $\textbf{[0.07, 0.25]}$ &  $\textbf{[0.23, 0.48]}$ \\
CW Variance &                        &  $\textbf{[0.02, 0.16]}$ &  $\textbf{[0.84, 0.98]}$ \\
1vA Entropy  &                        &                         &  $\textbf{[0.84, 0.98]}$ \\
\hline
\end{tabular}

    \subcaption*{Vessel Wall}
    \begin{tabular}{p{0.167\textwidth}|p{0.167\textwidth}|p{0.167\textwidth}|p{0.167\textwidth}p{0.167\textwidth}}
\hline
{} &     CW Variance &      1vA Entropy &           1vA MI \\
\hline
CW Entropy  &  $\textbf{[0.18, 0.42]}$ &  $\textbf{[0.01, 0.13]}$ &  $\textbf{[0.22, 0.46]}$ \\
CW Variance &                         &    $\textbf{[0.00, 0.10]}$ &    $\textbf{[0.90, 1.00]}$ \\
1vA Entropy  &                         &                         &    $\textbf{[0.90, 1.00]}$ \\
\hline
\end{tabular}

    \subcaption*{Lumen}
    \begin{tabular}{p{0.167\textwidth}|p{0.167\textwidth}|p{0.167\textwidth}|p{0.167\textwidth}p{0.167\textwidth}}
\hline
{} & CW Variance &      1vA Entropy &           1vA MI \\
\hline
CW Entropy  &      $[0.35, 0.61]$ &  $\textbf{[0.22, 0.46]}$ &           $[0.40, 0.67]$ \\
CW Variance &                     &  $\textbf{[0.02, 0.16]}$ &  $\textbf{[0.77, 0.95]}$ \\
1vA Entropy  &                     &                         &  $\textbf{[0.82, 0.97]}$ \\
\hline
\end{tabular}

\end{table}

\begin{table}[H]
\caption{95\% credible interval of the posterior distribution of patient-wise $p_{A > B}$ of the 10 best models for the BRATS-UNC performance measure per class and the class-specific uncertainty maps pairs A and B. The rows correspond to the uncertainty maps A and the columns to the uncertainty map B. The statistically significant results are reported in bold ($0.5 \notin I_{95\%}$). $p_{A > B} > 0.5$ means that uncertainty map A is better than uncertainty map B and the contrary if $p_{A > B} < 0.5$ (CW = Class-wise, 1vA = One versus all)}
\label{tab:top10classwisesignificance}
\centering
\subcaption*{Background}
\begin{tabular}{p{0.167\textwidth}|p{0.167\textwidth}|p{0.167\textwidth}|p{0.167\textwidth}p{0.167\textwidth}}
\hline
{} &    CW Variance &     1vA Entropy &          1vA MI \\
\hline
CW Entropy  &  $\textbf{[0.93, 1.00]}$ &  $\textbf{[0.93, 1.00]}$ &  $\textbf{[0.93, 1.00]}$ \\
CW Variance &                        &  $\textbf{[0.93, 1.00]}$ &  $\textbf{[0.93, 1.00]}$ \\
1vA Entropy  &                        &                        &  $\textbf{[0.00, 0.07]}$ \\
\hline
\end{tabular}

\subcaption*{Vessel Wall}
\begin{tabular}{p{0.167\textwidth}|p{0.167\textwidth}|p{0.167\textwidth}|p{0.167\textwidth}p{0.167\textwidth}}
\hline
{} &     CW Variance &     1vA Entropy &          1vA MI \\
\hline
CW Entropy  &  $\textbf{[0.82, 0.97]}$ &  $\textbf{[0.93, 1.00]}$ &  $\textbf{[0.93, 1.00]}$ \\
CW Variance &                         &  $\textbf{[0.93, 1.00]}$ &  $\textbf{[0.93, 1.00]}$ \\
1vA Entropy  &                         &                        &  $\textbf{[0.00, 0.07]}$ \\
\hline
\end{tabular}

\subcaption*{Lumen}
\begin{tabular}{p{0.167\textwidth}|p{0.167\textwidth}|p{0.167\textwidth}|p{0.167\textwidth}p{0.167\textwidth}}
\hline
{} &   CW Variance &     1vA Entropy &          1vA MI \\
\hline
CW Entropy  &  $\textbf{[0.90, 1.00]}$ &  $\textbf{[0.93, 1.00]}$ &  $\textbf{[0.93, 1.00]}$ \\
CW Variance &                       &  $\textbf{[0.93, 1.00]}$ &  $\textbf{[0.93, 1.00]}$ \\
1vA Entropy  &                       &                        &  $\textbf{[0.00, 0.07]}$ \\
\hline
\end{tabular}

\end{table}

\end{document}